%% file: main.tex
\documentclass[10pt,twocolumn,letterpaper]{article}

\usepackage[pagenumbers]{cvpr} % To force page numbers, e.g. for an arXiv version

\input{preamble}

\definecolor{cvprblue}{rgb}{0.21,0.49,0.74}
\usepackage[pagebackref,breaklinks,colorlinks,citecolor=cvprblue]{hyperref}
\usepackage{algorithm}
\usepackage{algorithmic}

\def\paperID{12779} % *** Enter the Paper ID here
\def\confName{CVPR}
\def\confYear{2024}

\title{Anomaly Detection with Conditioned Denoising Diffusion Models}

\author{Arian Mousakhan$^1$ \quad Thomas Brox$^1$ \quad Jawad Tayyub$^2$\\
    $^1$University of Freiburg \quad $^2$Endress+Hauser\\
    \texttt{\{mousakha,brox\}@cs.uni-freiburg.de}\\
    \texttt{jawad.tayyub@endress.com}
}

\begin{document}
\maketitle
\input{sec/0_abstract}    
\input{sec/1_intro}

\input{sec/2_related}
\input{sec/3_background}
\input{sec/4_method}

\input{sec/5_experiments}
\input{sec/6_conclusion}
{
    \small
    \bibliographystyle{ieeenat_fullname}
    \bibliography{main}
}

\input{sec/X_suppl}

\end{document}

%% file: preamble.tex
\usepackage[dvipsnames]{xcolor}

%% file: sec/0_abstract.tex
\begin{abstract}
Traditional reconstruction-based methods have struggled to achieve competitive performance in anomaly detection. In this paper, we introduce Denoising Diffusion Anomaly Detection (DDAD), a novel denoising process for image reconstruction conditioned on a target image. This ensures a coherent restoration that closely resembles the target image. Our anomaly detection framework employs the conditioning mechanism, where the target image is set as the input image to guide the denoising process, leading to a defectless reconstruction while maintaining nominal patterns. Anomalies are then localised via a pixel-wise and feature-wise comparison of the input and reconstructed image. Finally, to enhance the effectiveness of the feature-wise comparison, we introduce a domain adaptation method that utilises nearly identical generated examples from our conditioned denoising process to fine-tune the pretrained feature extractor. The veracity of DDAD is demonstrated on various datasets including MVTec and VisA benchmarks, achieving state-of-the-art results of \(99.8 \%\) and \(98.9 \%\) image-level AUROC respectively. Source code is available at \href{https://github.com/arimousa/DDAD}{GitHub}.  %All code and checkpoints will be made public upon acceptance.
\end{abstract}

%% file: sec/1_intro.tex
\section{Introduction}
\label{sec:intro}

Anomaly detection involves the identification and localisation of instances in data that are inconsistent with nominal observations. Detecting out-of-distribution data is a pivotal task in many fields of industry \cite{bergmann2019mvtec, zou2022spot}, medicine \cite{mood_2022_6362313, irvin2019chexpert} and video surveillance \cite{liu2018future}. In a supervised setting, a model is trained on a dataset with normal and abnormal examples. However, anomalies are usually unforeseen and these models often struggle during inference. Conversely, unsupervised methods model the distribution of only nominal samples to detect anomalies as patterns that deviate from the nominal distribution. Thus, they are not restricted to a finite set of anomalies.

\begin{figure*}[t]
    \centering
    \includegraphics[width=\textwidth]{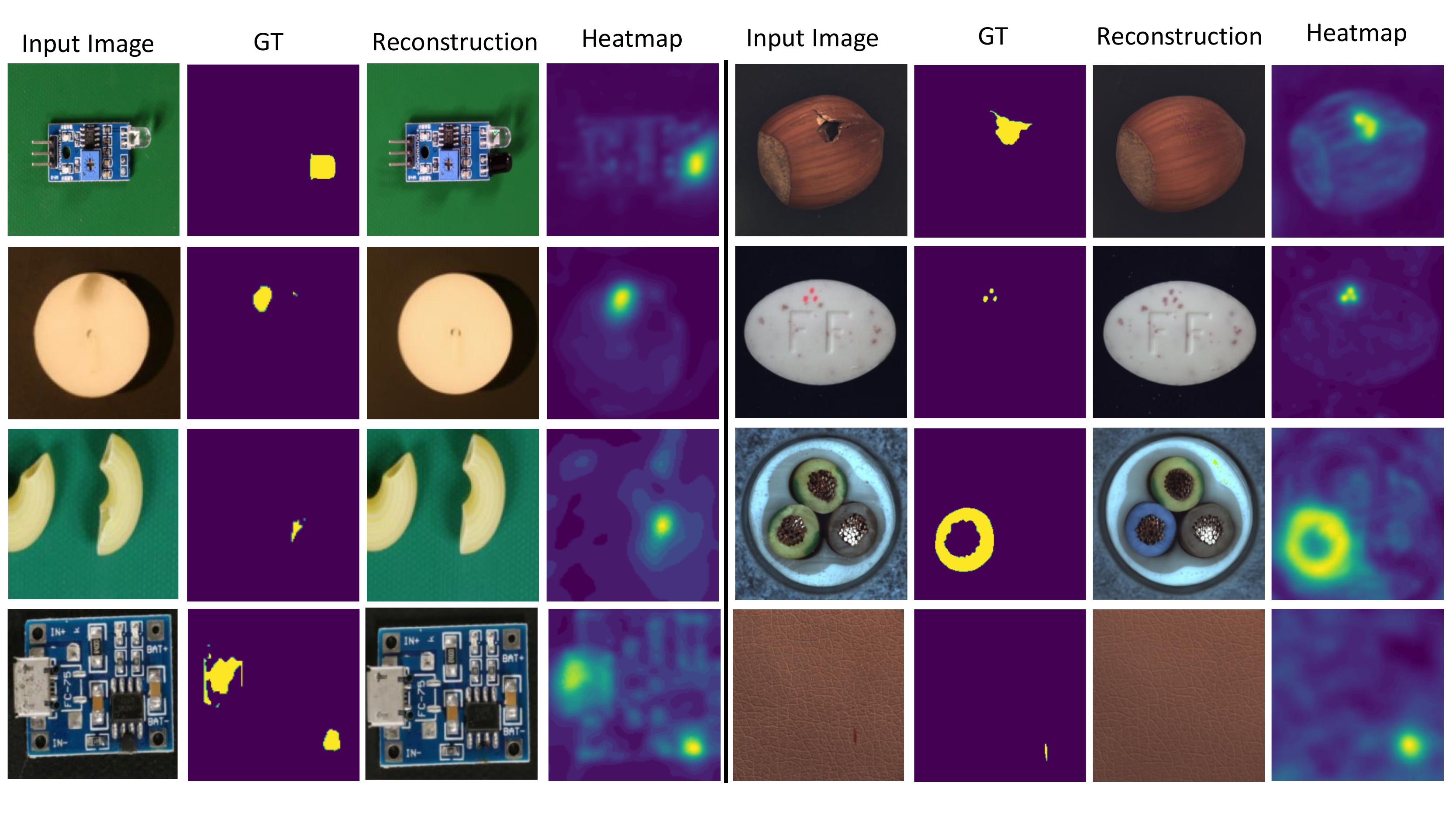}
    \caption{Our approach achieves defect-free reconstruction of input images that are devoid of anomalies. An accurate anomaly detection heatmap is computed. Note that reconstructions are analogous to the expected nominal approximation of the input. In the category of cables, an incorrectly placed green cable has been corrected to a blue one by the model. Such \textit{corrected images} may offer further benefit for the industry in repairing defects or worker training.}
    \label{fig:Teaser}
\end{figure*}

Representation-based methods \cite{roth2022towards, gudovskiy2022cflow, cohen2020sub,defard2021padim, deng2022anomaly,yu2021fastflow} rely on extracted features from pretrained neural networks to define the similarity metric for nominal samples and to approach the problem on a nearest neighbour strategy. Reconstruction-based methods \cite{akcay2019ganomaly,dehaene2020anomaly,kingma2013auto} learn a generative model from \textit{only} nominal training examples. Such models learn the entire distribution of nominal samples but are incapable of generating samples that deviate from this distribution. This allows for the detection of anomalies by comparing anomalous input with its predicted \textit{anomaly-free} reconstruction. However, past methods have suffered from inferior reconstruction quality or insufficient coverage of the nominal distribution, both resulting in erroneous comparisons between the reconstruction and the input image.

Recently, diffusion models \cite{sohl2015deep, ho2020denoising} have gained popularity as prolific deep generative models. This paper revisits reconstruction-based anomaly detection framework, harnessing the potential of diffusion models to generate an impressive reconstruction of anomalous images, see Figure \ref{fig:Teaser}. In this paper, we show that plain diffusion models are inapplicable to the anomaly detection task. Thus, we make the following contributions. First, we propose a conditioning mechanism that guides the denoising process to amend each perturbed image until it approximates a target image. This conditioning mechanism increases Image AUROC from \(85.7\%\) to \(92.4\%\) and from \(87.0\%\) to \(94.1\%\) on MVTec \cite{bergmann2019mvtec} and VisA \cite{zou2022spot}, respectively. Second, we discover that a combination of a pixel-wise and feature-wise comparison of the reconstruction and the input image boosts the detection and localisation precision. Third, we introduce an unsupervised domain adaptation technique to shift the domain of a pretrained feature extractor to the problem at hand. For this purpose, a similar image to a target image is generated by our denoising pipeline. The pretrained feature extractor is then fine-tuned by minimising the extracted features' distance from the two images. In order to avoid catastrophic forgetting of the pretrained network, we additionally include a distillation loss from a frozen feature extractor. Our domain adaptation technique instils invariance to nominal changes during reconstruction while preserving generality and learning the new domain. This domain-adapted feature comparison further lifts results to an Image AUROC of \(99.8 \%\) and \(98.9 \%\) on MVTec and VisA, surpassing not only reconstruction-based methods but state-of-the-art (SOTA) representation-based models. We additionally introduce a compressed version of \textit{DDAD}, denoted as \textit{DDAD-S}, tailored for applications constrained by limited resources.

%% file: sec/2_related.tex
\section{Related Work}
\label{sec:related}

\paragraph{Representation-based methods} Self-supervised learning has been used in the past to learn image features \cite{mathieu2015deep,noroozi2016unsupervised, gidaris2018unsupervised}, often by solving auxiliary tasks. In anomaly detection,  \cite{NEURIPS2018_5e62d03a, hendrycks2019using} have demonstrated that high-quality features facilitate the detection of anomalous samples. DN2 \cite{bergman2020deep} has successfully employed simple ResNets \cite{he2016deep}, pretrained on Imagenet \cite{russakovsky2015imagenet}, to extract informative features. Recent approaches such as SPADE \cite{cohen2020sub} uses a memory bank of nominal extracted features, PaDiM \cite{defard2021padim} uses locally constrained bag-of-features, PatchCore \cite{roth2022towards} uses a memory bank and neighborhood-aware patch-level features, CFLOW and FastFlow \cite{gudovskiy2022cflow, yu2021fastflow} use normalizing flow \cite{dinh2017density, kingma2018glow}, and US and RD4AD \cite{bergmann2020uninformed, deng2022anomaly} use a knowledge distillation method \cite{hinton2015distilling} for anomaly detection. All rely on pretrained feature extractors without any adaptation to the domain of the current problem. These models may fail when a pretrained feature extractor cannot provide informative features.
In this work, we utilise locally aware patch features, as proposed by \cite{roth2022towards}, to improve the comparison of the input image and its reconstruction at inference time. We propose a method to transfer knowledge of the current domain of feature extractors used in the aforementioned models, achieving superior performance.

\paragraph{Reconstruction-based methods} The initial frameworks for anomaly detection were developed based on the foundational concept that a generative model, trained on nominal samples, learns to accurately reconstruct nominal data while failing to reconstruct anomalies. Anomalous data typically deviate significantly from learned patterns leading to a poor reconstruction of anomalies at inference time. An early work \cite{lu2018anomaly} applied Variational Autoencoder (VAE) \cite{kingma2013auto} to detect anomalies in skin disease images. However, reconstructions were blurry and anomalies weren't adequately removed. Various techniques have since been proposed, \cite{bergmann2018improving} use a perceptual loss based on structural similarity (SSIM) to improve learning. \cite{sabokrou2018adversarially} deploy one generative model as a novelty detector connected end-to-end to a second network enhancing the inlier samples and distorted outliers. \cite{pidhorskyi2018generative} use an adversarial autoencoder to effectively compute the likelihood of a sample generated by the inlier distribution. However, these methods are only capable of one-class classification and do not localise anomalies. Ganomaly \cite{akcay2019ganomaly} makes use of a conditional GAN \cite{goodfellow2020generative,mirza2014conditional}, outperforming previous state-of-the-art models. \cite{ristea2022self, zavrtanik2021draem} use a discriminative end-to-end trainable surface anomaly paradigm for the detection and localisation of anomalies. These models rely on synthetic anomalies for training. Recently, denoising diffusion models have gained popularity for image, and audio generation \cite{sohl2015deep, ho2020denoising}. In the medical domain, denoising diffusion models have been used to detect brain tumours \cite{wolleb2022diffusion}. AnoDDPM \cite{wyatt2022anoddpm} showed that these models outperform GANs for anomaly detection in the medical domain. 

%% file: sec/3_background.tex
\section{Background}
\label{sec:background}
Denoising diffusion models \cite{sohl2015deep, ho2020denoising} are generative models, inspired by non-equilibrium thermodynamics, aiming to learn a distribution \(p_{\theta}(\mathbf{x})\) that closely resembles the data distribution \(q(\mathbf{x})\). Diffusion models generate latent noisy variables \(\mathbf{x}_1,...,\mathbf{x}_T\), having the same dimensions as the input data \(\mathbf{x}\sim q(\mathbf{x})\), by gradually adding noise \(\epsilon \sim \mathcal{N}(0,\mathbf{I})\) at each time step \(t\). This results in \(\mathbf{x}_T\) being complete noise normally distributed with mean 0 and variance 1. Given a pre-defined variance schedule $\beta_1 < \beta_2 < ... < \beta_T$ where \(\beta_t \in (0,1)\), the forward process over a series of $T$ steps is defined as follows:
\begin{equation} 
\label{forward process}
\begin{aligned}
    q(\mathbf{x}_{1:T}|\mathbf{x}) &= \prod_{(t\geq 1)}q(\mathbf{x}_t|\mathbf{x}_{t-1}), \\
    q(\mathbf{x}_t|\mathbf{x}_{t-1}) &= \mathcal{N}(\mathbf{x}_t;\sqrt{1-\beta_t}\mathbf{x}_{t-1},\beta_t\mathbf{I}).
\end{aligned}
\end{equation}
Given the additivity property, merging multiple Gaussians results in a Gaussian distribution. Therefore \( \mathbf{x}_t\) is directly computed at any arbitrary time step \(t\) by perturbing the input image \(\mathbf{x}\) as \(q(\mathbf{x}_t|\mathbf{x}) = \mathcal{N}(\mathbf{x}_t;\sqrt{\alpha_t}\mathbf{x},(1-\alpha_t) \mathbf{I})\), where \(\alpha_t = \prod_{i=1}^{t}(1-\beta_i)\). Despite the ease with which noise is introduced to an image, undoing this perturbation is inherently challenging. This is referred to as \textit{reverse} or \textit{denoising process} in DDPM \cite{NEURIPS2020_4c5bcfec} defined by a parameterised function \(p_{\theta}(\mathbf{x}_{t-1}|\mathbf{x}_t) = \mathcal{N}(\mathbf{x}_{t-1};\mu_{\theta}(\mathbf{x}_t,t),\beta_t\mathbf{I} )\), where the mean is derived using the learnable function \(\epsilon_{\theta}^{(t)}(\mathbf{x}_t)\). DDPM suggests the training objective \(||\epsilon_{\theta}^{(t)}(\mathbf{x}_t)-\epsilon||^2\) to train the model.

Denoising Diffusion Implicit Models (DDIM) \cite{song2021denoising} accelerate upon DDPM by employing a non-Markovian sampling process. DDIM uses an implicit density model rather than an explicit one used in DDPM. DDIM suggests a sampling process \(q_{\sigma}(\mathbf{x}_{t-1}|\mathbf{x}_t,\mathbf{x})\) by defining a new variance schedule. Based on \(\mathbf{x}_t = \sqrt{\alpha_t} \mathbf{x}+\sqrt{1-\alpha_t}\epsilon\), one can predict the denoised observation \(\mathbf{x}_0\) as follows: \begin{equation}
    \label{prediction of x_zero}
    f_{\theta}^{(t)}(\mathbf{x}_t) := (\mathbf{x}_t - \sqrt{1-\alpha_t}.\epsilon_{\theta}^{(t)}(\mathbf{x}_t))/\sqrt{\alpha_t}. 
\end{equation}
Having defined the generative process \(p_{\theta}^{(t)}(\mathbf{x}_{t-1}|\mathbf{x}_t) = q_{\sigma}(\mathbf{x}_{t-1}|\mathbf{x}_t, f_{\theta}^{(t)}(\mathbf{x}_t))\), accordingly via 
\begin{equation}
\mathbf{x}_{t-1} = \sqrt{\alpha_{t-1}} f_{\theta}^{(t)}(\mathbf{x}_t) + \sqrt{1-\alpha_{t-1}-\sigma_t^2}.\epsilon_{\theta}^{(t)}(\mathbf{x}_t) + \sigma_t\epsilon_t, 
\end{equation}
where \(\sigma_t\) determines the stochasticity of the sampling process, one can generate new samples.

The connection between diffusion models and score matching \cite{song2019generative} was introduced by \cite{song2021scorebased} and derived a score-based function to estimate the deviation that should happen at each time step to make a less noisy image. It can be written as:
\begin{equation}
\label{unconditional score function}
    \nabla_{\mathbf{x}_t} \log p_{\theta}(\mathbf{x}_t) = - \frac{1}{\sqrt{1-\alpha}} \epsilon_{\theta}^{(t)}(\mathbf{x}_t),
\end{equation}
which \cite{dhariwal2021diffusion} used this property to introduce a classifier guidance mechanism. Similarly, we leverage the score-based function to introduce our conditioned denoising process in the following section. Note that, in this paper, we refer to \(\mathbf{x}\) as the input image and \(\mathbf{x}_0\) as its reconstruction.

%% file: sec/4_method.tex
\section{Method}
\label{sec:method}

\begin{figure*}[t]
    \centering
    \includegraphics[width=\textwidth]{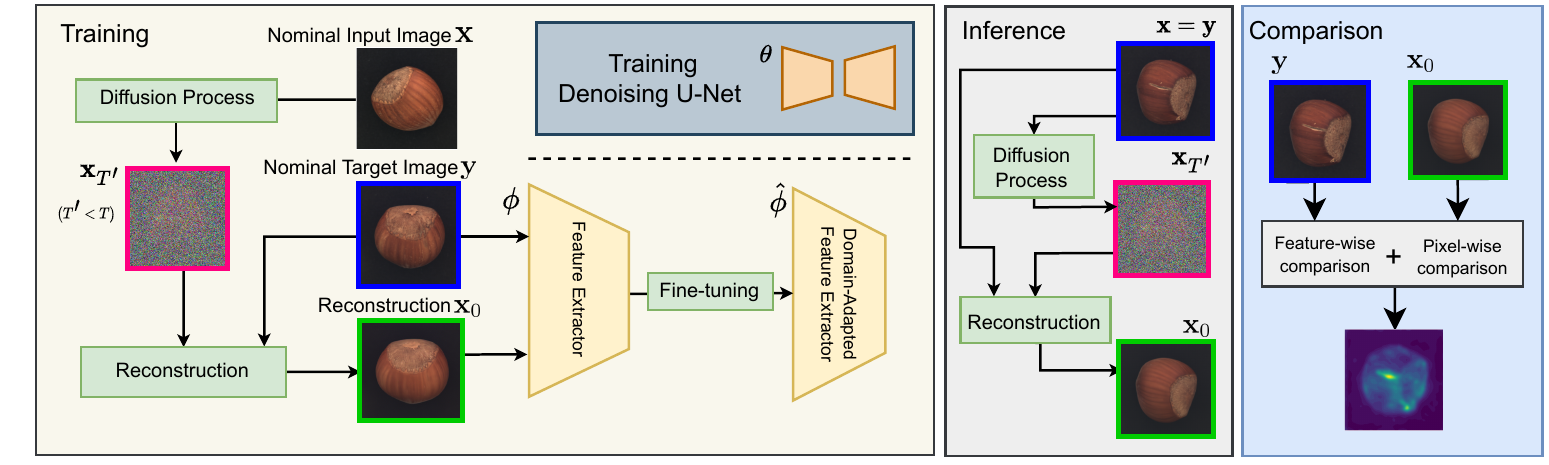}
    \caption{Framework of \textit{DDAD}. After a denoising U-Net has been trained, the feature extractor is adapted to the problem domain by minimising the distance between the extracted features of a target image and a generated image which resembles the target image. At inference time, after perturbing the input image, the denoising process is conditioned on the same input image to make an anomaly-free reconstruction. Finally, the reconstructed image is compared with the input through both pixel and feature matching to generate an accurate anomaly localisation.}
    \label{fig:Framework}
\end{figure*}

In this section, we detail our DDAD framework. We first present our proposed conditioning mechanism for reconstruction. We then explain how it is used to eradicate anomalies while preserving nominal information. We then present a robust approach to compare the reconstructed image with the input, resulting in an accurate anomaly localisation. An overview of \textit{DDAD} is presented in Figure \ref{fig:Framework}.

\subsection{Conditioned Denoising Process for Reconstruction}
\label{Conditional Reverse Noising Process for Reconstruction}
Given a target image \(\mathbf{y}\) and a perturbed image \(\mathbf{x}_t\), our aim is to denoise \(\mathbf{x}_t\) step-by-step to result in an image starkly similar to \(\mathbf{y}\). To this end, we condition the score function on the target image to achieve a posterior score function \(\nabla_{\mathbf{x}_t} \log p_{\theta}(\mathbf{x}_t|\mathbf{y})\). However, directly calculating this posterior score function is challenging, since \(\mathbf{x}_t\) and \(\mathbf{y}\) do not consist of the same signal-to-noise ratio. To tackle this challenge, we rely on the assumption that if the reconstructed image \(\mathbf{x}_0\) is similar to \(\mathbf{y}\), therefore, adding the same noise as \(\mathbf{x}_t\) consists of, to the \(\mathbf{y}\), would result in \(\mathbf{x}_t \sim \mathbf{y}_t\). This helps to guide \(\mathbf{x}_t\) towards \(\mathbf{y}_t\) at each denoising step.

In order to compute \(\mathbf{y}_t\), we add \(\epsilon_{\theta}^{(t)}(\mathbf{x}_t)\) which is predicted by the trained diffusion model, to \(\mathbf{y}\). Following this, the condition is modified by replacing \(\mathbf{y}\) by \(\mathbf{y}_t\), resulting in \(\nabla_{\mathbf{x}_t} \log p_{\theta}(\mathbf{x}_t|\mathbf{y}_t)\) to guide the denoising process. Based on Bayes' rule, this decomposes as follows:
\begin{equation}
  \label{posterior distribution}
    \nabla_{\mathbf{x}_t} \log p_{\theta}(\mathbf{x}_t|\mathbf{y}_t)=\nabla_{\mathbf{x}_t} \log p_{\theta}(\mathbf{x}_t) + \nabla_{\mathbf{x}_t} \log p_{\theta}(\mathbf{y}_t|\mathbf{x}_t).
\end{equation}
The unconditional score term \(\nabla_{\mathbf{x}_t} \log p_{\theta}(\mathbf{x}_t)\) can be directly calculated from Eq. \ref{unconditional score function}. In many cases calculating the conditional score (or likelihood) \(\nabla_{\mathbf{x}_t} \log p_{\theta}(\mathbf{y}_t|\mathbf{x}_t)\) is intractable. Nevertheless, having calculated \(\mathbf{y}_t\) allows for directly computing this likelihood. Intuitively, the likelihood \( \nabla_{\mathbf{x}_t} \log p_{\theta}(\mathbf{y}_t|\mathbf{x}_t) \) can be viewed as a correction score for a deviation that occurs in \(\mathbf{x}_t\) from \(\mathbf{y}_t\) at each denoising step. Knowing that both \(\mathbf{x}_t\) and \(\mathbf{y}_t\) consist of the same noise, this deviation is only present at the image (signal) level. Consequently, the divergence can be calculated by \(\mathbf{y}_t - \mathbf{x}_t\), and an adjusted noise term \(\hat{\epsilon}\) is updated as follows:
\begin{equation}
  \label{modified noise}
  \hat{\epsilon} = \epsilon_{\theta}^{(t)}(\mathbf{x}_t) - w \sqrt{1-\alpha_t} (\mathbf{y}_t- \mathbf{x}_t),
\end{equation}
where \(w\) controls the power of the conditioning. Given \(\hat{\epsilon}\), the new prediction \(\hat{f}_{\theta}^{(t)}(\mathbf{x}_t)\) is calculated using Eq. \ref{prediction of x_zero}. 

Finally, the less-noisy image \(\mathbf{x}_{t-1}\) is calculated via the denoising process as follows:
\begin{equation}
  \mathbf{x}_{t-1} = \sqrt{\alpha_{t-1}} \hat{f}_{\theta}^{(t)}(\mathbf{x}_t) + \sqrt{1-\alpha_{t-1}-  \sigma_t^2} \hat{\epsilon} + \sigma_t\epsilon_t.
\end{equation}
The summary of our reconstruction process is shown in Algorithm \ref{algorithm1}.

\begin{algorithm}[ht]
    \centering
    \caption{Reconstruction Process}\label{algorithm1}
    \begin{algorithmic}[1]
        \STATE \(\mathbf{x}_{T^{\prime}} \leftarrow \sqrt{\alpha_{T^{\prime}}} \mathbf{x} + \sqrt{1-\alpha_{T^{\prime}}} \epsilon_t \)
        \FOR {\textbf{all} \( t = T^{\prime},...,1\)}
        \STATE \( \mathbf{y}_t \leftarrow \sqrt{\alpha_t} \mathbf{y} + \sqrt{1-\alpha_t} \epsilon_{\theta}^{(t)}(\mathbf{x}_t) \)
        \STATE \( \hat{\epsilon} \leftarrow \epsilon_{\theta}^{(t)}(\mathbf{x}_t) - w \sqrt{1-\alpha_t} (\mathbf{y}_t - \mathbf{x}_t)\)
        \STATE \( \hat{f}_{\theta}^{(t)}(\mathbf{x}_t) \leftarrow (\mathbf{x}_t - \sqrt{1-\alpha_t}.\hat{\epsilon})/\sqrt{\alpha_t} \)
        \STATE \( \mathbf{x}_{t-1} \leftarrow \sqrt{\alpha_{t-1}} \hat{f}_{\theta}^{(t)}(\mathbf{x}_t) \)
        \STATE \; \; \; \;  \; \; \(+ \sqrt{1- \alpha _{t-1}-\sigma_t^2} \hat{\epsilon} + \sigma_t \epsilon_t\)
        \ENDFOR
        \RETURN \(\mathbf{x}_0\)
    \end{algorithmic}
\end{algorithm}

\subsection{Reconstruction for Anomaly Detection}
\label{Reconstruction for Anomaly Detection}
For anomaly detection tasks, the target image \(\mathbf{y}\) is set as the input image \(\mathbf{x}\). This enables the denoising process, which is conditioned on \(\mathbf{y}\), to generate an anomaly-free approximation of \(\mathbf{x}\). Since the model is only trained on nominal data, anomalous regions lie in the low probability density of \(p_{\theta}(\mathbf{x})\). Therefore, during denoising, the reconstruction of anomalies falls behind the nominal part. 

Over an entire trajectory, earlier steps focus on the abstract picture of the image whereas later steps aim to reconstruct fine-grained details. Since anomalies mostly emerge at a fine level, the starting denoising time step can be set earlier than complete noise i.e. \(T^{\prime} < T\), where a sufficient amount of signal-to-noise ratio is present. Note that the model is trained on complete trajectories. 

We label our model as \textit{DDAD-n}, where \textit{n} refers to the number of denoising iterations.

\subsection{Anomaly Scoring}
\label{Anomaly Scoring}
In the simplest case, we can detect and localise anomalies via a pixel-wise comparison between the input and its reconstruction. However, comparing only pixel distances of two images may not capture all anomalies such as poked parts or dents, whereby visible colour variations are not present. Therefore, we additionally compute distances between image features extracted by deep neural networks to also capture perceptual similarity \cite{zhang2018unreasonable, dosovitskiy2016generating}. Features are sensitive to changes in edges and textures where a pixel-wise comparison may fail, but they are often robust against slight transformations. We discovered that employing both image and feature level comparisons yields the most precise anomaly localisation.

% In the following subsection, we present an unsupervised domain adaptation technique designed to enhance the feature extractor's responsiveness to domain-specific changes. Nevertheless,

Given a reconstructed image \(\mathbf{x}_0\) and the target image \(\mathbf{y}\), we define a pixel-wise distance function \(D_p\) and a feature-wise distance function \(D_f\) to derive the anomaly heatmap. \(D_p\) is calculated based on the \(\mathcal{L}_1\) norm in pixel space. At the feature level, similar to PatchCore \cite{roth2022towards} and PaDiM \cite{defard2021padim}, we utilise adaptive average pooling to spatially smooth each individual feature map. Features within a given patch are aggregated in a single representation, resulting in the same dimensionality as the input feature. Finally, a cosine similarity is utilised to define \(D_f\) as:
\begin{equation}
\label{feature distance}
    D_f(\mathbf{x}_0,\mathbf{y}) = \sum_{j \in J} \left(1-\cos(\phi_j(\mathbf{x}_0),\phi_j(\mathbf{y}))\right),
\end{equation}

% \begin{equation}
% \label{cosine distance}
%     D_f(\mathbf{x}_0,\mathbf{y}) = \sum_{j \in J} \left(1-\frac{\phi_j(\mathbf{x}_0)^T.\phi_j(\mathbf{y})}{\|\phi_j(\mathbf{x}_0)\| \|\phi_j(\mathbf{y})\|}\right).
% \end{equation}
where \(\phi\) \cite{zagoruyko2016wide, he2016deep} refers to a pretrained feature extractor and \(j \in J\) is the set of layers considered. We only use \(j \in \{2,3\}\) to retain the generality of the used features \cite{roth2022towards}. Finally, we normalise the pixel-wise distance \(D_p\) to share the same upper bound as the feature-wise distance \(D_f\). Consequently, the final anomaly score function is a combination of the pixel and the feature distance:
\begin{equation}
    D_{anomaly} = \left(v \frac{\max(D_f)}{\max(D_p)}\right) D_p +  D_f,
\end{equation}
where \(v\) controls the importance of the pixel-wise distance. 

% \begin{algorithm}[ht]
%     \centering
%     \caption{Anomaly Scoring}\label{algorithm2}
%     \begin{algorithmic}[1]
%                 \STATE \(D_p \leftarrow \mathcal{L}_1(\mathbf{x}_0,\mathbf{y})\)
%         \STATE \(D_f \leftarrow 0\)
%         \FOR{\textbf{all} \(j \in J\) }
%             \STATE \(D_f \leftarrow D_f + \left(1-\cos(\phi_j(\mathbf{x}_0),\phi_j(\mathbf{y}))\right)\)
%         \ENDFOR
%     \STATE \( D_p \leftarrow \left(\frac{\max(D_f)}{\max(D_p)}\right) D_p \)
%     \STATE \( D_{anomaly} \leftarrow v D_p +  D_f\)
%     \RETURN \(D_{anomaly}\)
%     \end{algorithmic}
% \end{algorithm}

\subsection{Domain Adaptation}
\label{Domain Adaptation}
In Section \ref{Anomaly Scoring} we used a pretrained feature extractor for feature-wise comparison between an input image and its reconstruction. However, these networks are trained on ImageNet and do not adapt well to domain-specific characteristics of an anomaly detection task and a specific category. We propose a novel unsupervised domain adaptation technique by converging different extracted layers from two nearly identical images. This helps the networks become agnostic to nominal changes that may occur during reconstruction, at the same time learning the problem's domain. To achieve this, we first sample a random image \(\mathbf{x}\) from the training dataset and perturb it with noise to obtain \(\mathbf{x}_t\). Similarly, we randomly select a target image \(\mathbf{y}\) from the training dataset. Given a trained denoising model \(\theta\), a noisy image \(\mathbf{x}_t\) is denoised to \(\mathbf{x}_0\) to approximate \(\mathbf{y}\). Features are then extracted from the reconstructed and target image, denoted as \(\phi_j(\mathbf{x}_0)\) and \(\phi_j(\mathbf{y})\). With the assumption that \(\mathbf{x}_0 \sim \mathbf{y}\), their feature should be similar. Therefore, the network \(\phi\) is fine-tuned by minimising the distance between extracted features. A loss function \(\mathcal{L}_{Similarity}\), based on cosine similarity, is employed for each of the final activation layers of the \(j^{th}\) spatial resolution block. This transfers the pretrained model \(\phi\) to the domain-adapted network \(\hat{\phi}\). Nevertheless, we observe that the generalisation of the network diminishes after several iterations while learning the patterns of the new dataset. To mitigate this, we incorporate a distillation loss from a frozen feature extractor \(\overline{\phi}\) which mirrors the state of the network \(\phi\) prior to domain adaptation. This distillation loss safeguards the feature extractor from losing its generality during adaptation to the new domain. Consequently, the domain adaptation loss \(\mathcal{L}_{DA}\) can be expressed as follows:

% \begin{equation}
% \label{cosine distance}
%     L(\mathbf{x}_0,\mathbf{y}) = \sum_{j \in J} \left(1-\cos(\phi_j(\mathbf{x}_0),\phi_j(\mathbf{y}))\right) +  {\lambda}_{DL}(\sum_{j \in J} \left(1-\cos(\phi_j(\mathbf{y}),\overline{\phi}_j(\mathbf{y}))\right) + \sum_{j \in J} \left(1-\cos(\phi_j(\mathbf{x}_0),\overline{\phi}_j(\mathbf{x}_0))\right)).
% \end{equation}

\input{tables/mvtecAUROC}

\begin{equation}
\label{fine-tuning loss}
\begin{aligned}
    % \mathbb{E}(\phi) & = 
    \mathcal{L}_{DA} & = \mathcal{L}_{Similarity}(\mathbf{x_0},\mathbf{y}) + {\lambda}_{DL}\mathcal{L}_{DL}(\mathbf{x_0},\mathbf{y})  \\
     & = \sum_{j \in J} \left(1-\cos(\phi_j(\mathbf{x}_0),\phi_j(\mathbf{y}))\right) \\
     & + {\lambda}_{DL} \sum_{j \in J} \left(1-\cos(\phi_j(\mathbf{y}),\overline{\phi}_j(\mathbf{y}))\right) \\
     & + {\lambda}_{DL}\sum_{j \in J} \left(1-\cos(\phi_j(\mathbf{x}_0),\overline{\phi}_j(\mathbf{x}_0))\right),
\end{aligned}
\end{equation}

where \({\lambda}_{DL}\) determines the significance of distillation loss \(L_{DL}\). For our experiments, \(J\) is set as \(\{1,2,3\}\). The resulting feature extractor is resilient to slight changes during reconstruction. In Appendix, Section \ref{Robustness to anomalies on the background}, we highlight its role in making the model robust to nominal variation of the object and spurious anomalies in the background present in the reconstruction.

%% file: tables/mvtecAUROC.tex
\begin{table*}[ht]
\centering
\caption{A detailed comparison of Anomaly Classification and Localisation performance of various methods on MVTec benchmark \cite{bergmann2019mvtec} in the format of (image AUROC,pixel AUROC). The first five rows represent texture categories,and the next nine rows represent object categories.} 
\resizebox{\textwidth}{!}{\begin{tabular}{cccccccccc}
\toprule
  \multicolumn{1}{c}{}&\multicolumn{3}{c}{Representation-based}&\multicolumn{5}{c}{Reconstruction-based}\\
   \cmidrule(rl){2-4} \cmidrule(rl){5-10}
 \textbf{Method}  &RD4AD\cite{deng2022anomaly} &PatchCore\cite{roth2022towards} &SimpleNet \cite{liu2023simplenet} & GANomaly \cite{akcay2019ganomaly} & RIAD \cite{ZAVRTANIK2021107706} &Score-based PR \cite{Shin_2023_ICCV} & DRAEM \cite{zavrtanik2021draem}& DDAD-S-10 &DDAD-10 \\
 \midrule
 \textbf{Carpet}&(98.9,\textbf{98.9}) &  (98.7,\textbf{98.9}) &(\textbf{99.7},98.2)&(20.3,-) &(84.2,96.3) &(91.7,96.4) &(97.0,95.5)& (98.2,98.6)  & (99.3,98.7)\\
 \textbf{Grid} &(\textbf{100},99.3) &  (99.7,98.3) &(99.7,98.8)&(40.4,-)  &(99.6,98.8) &(100,98.9) &(99.9,\textbf{99.7})&(\textbf{100},98.4)&(\textbf{100},99.4)\\
 \textbf{Leather} &(\textbf{100},\textbf{99.4}) &(\textbf{100},99.3)&(\textbf{100},99.2) &(41.3,-)  &(100,99.4) &(99.9,99.3)  &(\textbf{100},98.6) & (\textbf{100},99.2)& (\textbf{100},\textbf{99.4})  \\
 \textbf{Tile} &(99.3,95.6) & (\textbf{100},\textbf{99.3}) &(99.8,97.0)&(40.8,-)  &(98.7,89.1) &(99.8,96.8)  &(99.6,99.2) & (\textbf{100},98.2)& (\textbf{100},98.2)\\
 \textbf{Wood} &(99.2,95.3) & (99.2,95.0) &(\textbf{100},94.5) &(74.4,-)  &(93.0,85.8) &(96.1,95.4)  &(99.1,\textbf{96.4}) & (99.9,95.1) & (\textbf{100},95.0)\\
  \midrule
 \textbf{Bottle} &(\textbf{100},98.7) &  (\textbf{100},98.6) &(\textbf{100},98.0) &(25.1,-)  &(99.9,98.4) &(100,95.9)  &(99.2,\textbf{99.1}) & (\textbf{100},98.5) & (\textbf{100},98.7)  \\
 \textbf{Cable}  &(95.0,97.4) &(99.5,\textbf{98.4})&(\textbf{99.9},97.6) &(45.7,-)  &(81.9,84.2) &(94.2,96.9)  &(91.8,94.7) & (99.8,98.3) & (99.4,98.1)\\
 \textbf{Capsule}  &(96.3,98.7)&(98.1,98.8)&(97.7,\textbf{98.9}) &(68.2,-)  &(88.4,92.8) &(97.2,96.6)  &(98.5,94.3)&(\textbf{99.4},96.0)   & (\textbf{99.4},95.7) \\
 \textbf{Hazelnut} &(99.9,98.9) & (\textbf{100},98.7)&(\textbf{100},97.9) &(53.7,-)  &(83.3,96.1) &(98.6,98.7)  &(\textbf{100},\textbf{99.7})& (99.8,98.4)& (\textbf{100},98.4) \\
 \textbf{Metal nut}&(\textbf{100},97.3) & (\textbf{100},98.4) &(\textbf{100},98.8) &(27.0,-)  &(88.5,92.5) &(96.6,96.6)  &(98.7,\textbf{99.5}) & (\textbf{100},98.1)  &(\textbf{100},99.0) \\
 \textbf{Pill}&(96.6,98.2) & (99.8,98.9)  &(99.0,98.6)&(47.2,-)  &(83.8,95.7) &(96.1,98.2) &(98.9,97.6) & (99.5,\textbf{99.1})&(\textbf{100},\textbf{99.1}) \\
 \textbf{Screw}  &(97.0,\textbf{99.6})&(98.1,99.4) &(98.2,99.3) &(23.1,-)  &(84.5,98.8) &(98.6,99.5)  &(93.9,97.6) & (98.3,99.0) & (\textbf{99.0},99.3) \\
 \textbf{Toothbrush} &(99.5,\textbf{99.1})&(\textbf{100},98.7) &(99.7,98.5)&(37.2,-)  &(100,98.9) &(98.1,97.8)  &(\textbf{100},98.1)& (\textbf{100},98.7) &(\textbf{100},98.7) \\
 \textbf{Transistor} &(96.7,92.5)&(\textbf{100},96.3)&(\textbf{100},\textbf{97.6}) &(44.0,-)  &(90.9,87.7) &(98.7,94.7)  &(93.1,90.9) & (\textbf{100},95.3) & (\textbf{100},95.3) \\
 \textbf{Zipper} &(98.5,98.2) & (99.4,98.8) &(99.9,\textbf{98.9}) &(43.4,-)  &(98.1,97.8) &(99.9,98.8)  &(\textbf{100},98.1) & (99.9,97.5) & (\textbf{100},98.2)\\
 \midrule
\textbf{Average} &(98.5,97.8) &(99.1,\textbf{98.1}) &(99.6,\textbf{98.1}) &(42.1,-)  &(91.7,94.2) &(97.7,97.4)  &(98.0,97.3) & (99.7,97.9) & (\textbf{99.8},\textbf{98.1})\\
\bottomrule
\end{tabular}}
\label{Table:MVTecAUROC}
\end{table*}

%% file: sec/5_experiments.tex
\section{Experiments}
\label{sec:experiments}
\subsection{Datasets and Evaluation Metrics}
We demonstrate the integrity of DDAD on three datasets: MVTec, VisA and MTD. Our model correctly classifies all samples in 11 out of 15 and 4 out of 12 categories in MVTec and VisA, respectively. The MVTec Anomaly Detection benchmark \cite{bergmann2019mvtec} is a widely known industrial dataset comprising 15 classes with 5 textures and 10 objects. Each category contains anomaly-free samples for training and various anomalous samples for testing ranging from small scratches to large missing components. We also evaluate our model on a new dataset called VisA \cite{zou2022spot}. This dataset is twice the size of MVTec comprising 9,621 normal and 1,200 anomalous high-resolution images. This dataset exhibits objects of complex structures placed in sporadic locations as well as multiple objects in one image. Anomalies include scratches, dents, colour spots, cracks, and structural defects. We also experimented on the Magnetic Tile Defects (MTD) dataset \cite{huang2020surface}. This dataset is a single-category dataset with 925 nominal training images and 5 sub-categories of different types of defects totalling 392 test images. We use \(80\%\) of defect-free images as the training set.

For MVTec and VisA datasets, we train the denoising network on images of size \( 256 \times 256 \) and, for comparison, images are cropped to \(224 \times 224\). No data augmentation is applied to any dataset, since augmentation transformations may masquerade as anomalies.

We assess the efficacy of our model by utilizing the Area Under Receiver Operator Characteristics (AUROC) metric, both at the image and pixel level. For image AUROC, we determine the maximum anomaly score across pixels and assign it as the overall anomaly score of the image. A one-class classification is then used to calculate the image AUROC for anomaly detection. For pixel level, in addition to pixel AUROC, we employ the Per Region Overlap (PRO) metric \cite{bergmann2020uninformed} for a more comprehensive evaluation of localisation performance. The PRO score treats anomaly regions of varying sizes equally, making it a more robust metric than pixel AUROC.

\input{tables/visaAUROC}

\input{tables/MTD}

\subsection{Experimental Setting}
To train our denoising model, we employ the modified UNet framework introduced in \cite{dhariwal2021diffusion}. For our compact model \textit{DDAD-S}, we reduced the base channels from 64 to 32 and the number of attention layers from 4 to 2. While \textit{DDAD} comprises 32 million parameters, \textit{DDAD-S} consists of only 8 million parameters. This reduction not only accelerates training and inference but also maintains comparable performance to our larger model. Consequently, \textit{DDAD-S} proves to be a more viable choice for edge devices within a resource-constrained production line. Complete implementation details are provided in Appendix, Section \ref{Implementation Details}. Furthermore, the selection of values of the two hyperparameters \(w\) and \(v\) are presented in Appendix, Section \ref{Hyperparameters}. Note that although the model is trained using \(T=1000\), we empirically identified \(T^{\prime} = 250\) as the optimal noise time step. This choice strikes a favourable balance between signal and noise in the context of our study.

\input{tables/MVTec-VisA-PRO}

\begin{figure}
    \centering
    \includegraphics[width=0.9\linewidth]{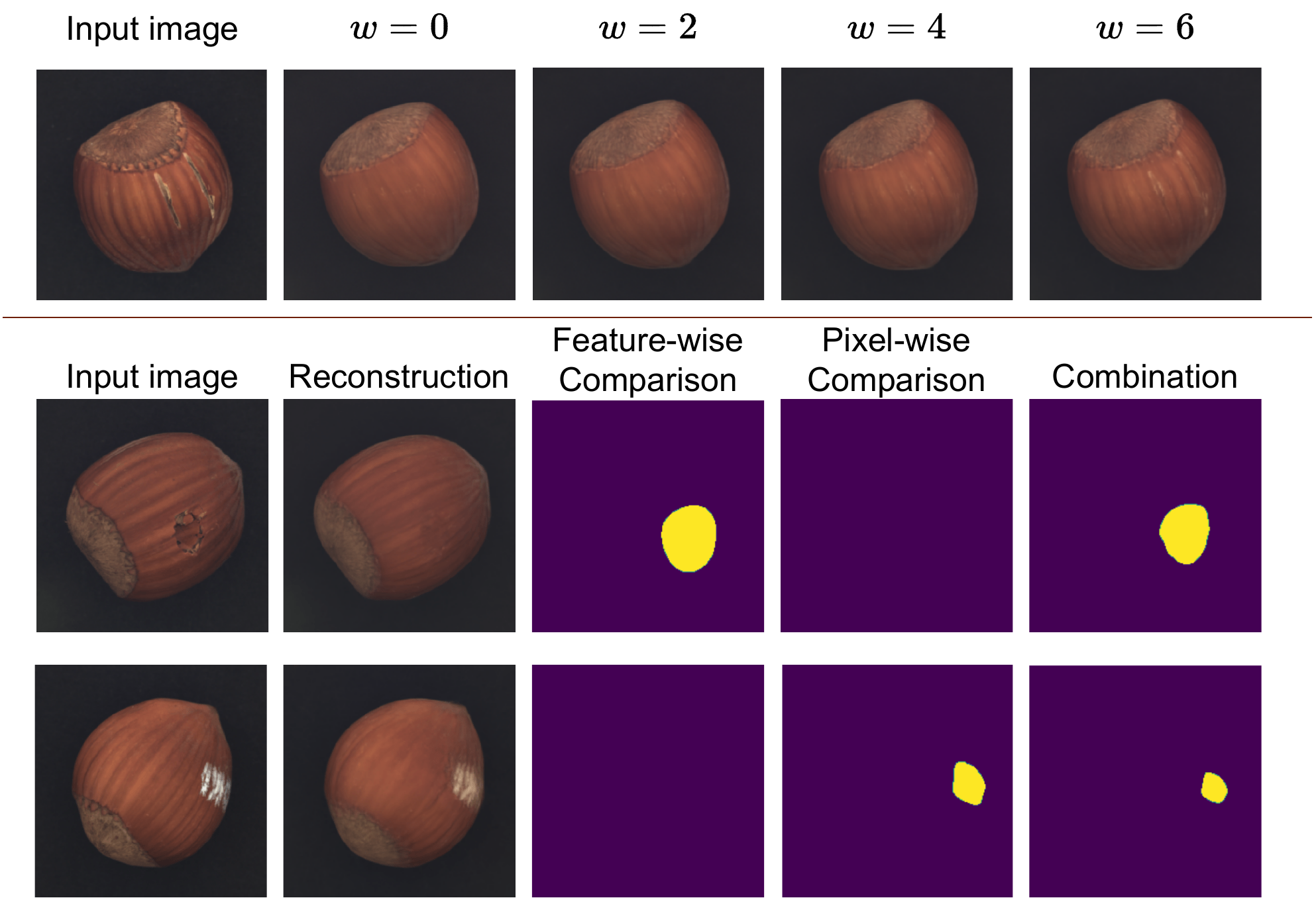}
    \caption{Top: Influence of conditioning parameter on reconstruction outcomes. Bottom: The first row illustrates a scenario where pixel-wise comparison proves ineffective, while the second row showcases a failure in feature-wise comparison. It is demonstrated that a combination leads to accurate detection in both cases.}
    \label{fig:Condition}
\end{figure}

\subsection{Experimental Results and Discussions}
\label{Experimental Results and Discussions}

Anomaly detection results on MVTec, VisA and MTD datasets are shown in Tables \ref{Table:MVTecAUROC}, \ref{Table: VisA AUROC}, and \ref{BTAD MTD AUROC} respectively. Our proposed framework \textit{DDAD} outperforms all existing approaches, not only the reconstruction-based but also representation-based methods, achieving the highest Image AUROC in all datasets. 
The proposed use of diffusion models not only enables anomaly detection and localisation but also the reconstruction of anomalies, based on generative modelling, which has been a longstanding idea, having limited success in anomaly detection.

\begin{figure*}
    \centering  
    \includegraphics[width=\linewidth]{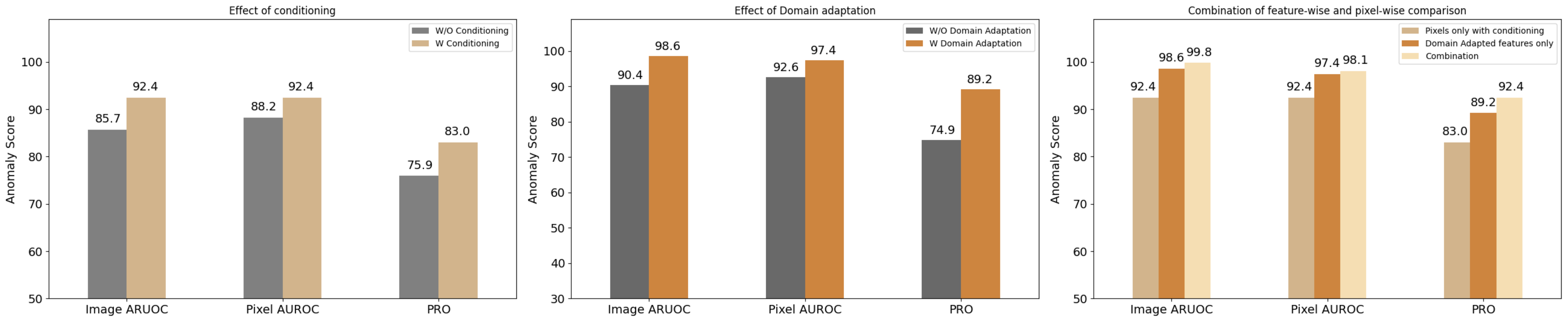}
    \caption{Effectiveness of various components of our model on anomaly detection and segmentation. Left: Effectiveness of conditioning based on only pixel-wise image comparison. Middle: Performance increase due to domain adaptation of feature extractor. The conditioning is applied for reconstruction. Right: Impact of merging feature-wise and pixel-wise image comparison. All results are shown on MVTec \cite{bergmann2019mvtec} dataset.}
    \label{fig:Comparison_methods_MVTec}
\end{figure*}

In Figure \ref{fig:Comparison_methods_MVTec}, we demonstrate the impact of each module of our framework on the MVTec dataset. Ablations with VisA are added to the Appendix, Section  \ref{Ablation on VisA}. We have shown 
plain diffusion models alone are not sufficient to lift reconstruction-based methods up to a competitive level. We have observed that applying the conditioning mechanism raises anomaly detection and localisation by 6.7\% and 4.2\%, respectively, in comparison to an unconditional denoising process, based on pixel-wise comparison. This demonstrates the ability of our guidance to increase the quality of reconstruction. Additionally, the use of diffusion-based domain adaptation adds \(8.2 \%\) and \(4.8 \%\) to the feature-wise comparison, and the combination of the pixel and feature level raises the final performance by \(1.2 \%\) and \(0.7 \%\) on anomaly detection and localisation respectively. Comprehensive analysis justification for the use of both pixel and feature comparisons is discussed in Appendix, Section \ref{Need for both pixel and feature comparison.}.

\textit{DDAD} performance on the PRO metric is presented in Table \ref{Table: MVTec-VisA PRO}. \textit{DDAD} achieves SOTA results on VisA and competitive results to PaDiM \cite{defard2021padim} and PatchCore \cite{roth2022towards} in MVTec. The inferior pixel-level performance compared to image-level performance can be attributed to the initial denoising point \(T^{\prime} = 250\), which presents a greater challenge to reconstruct large missing components (such as some samples in the transistor category). However, starting from earlier time steps introduces ambiguities in the reconstruction and leads to increased inference time. Some failure modes of the model are presented in the Appendix, Section \ref{Mislocalisation}.

\begin{figure*}
    \centering
    \includegraphics[width = \textwidth]{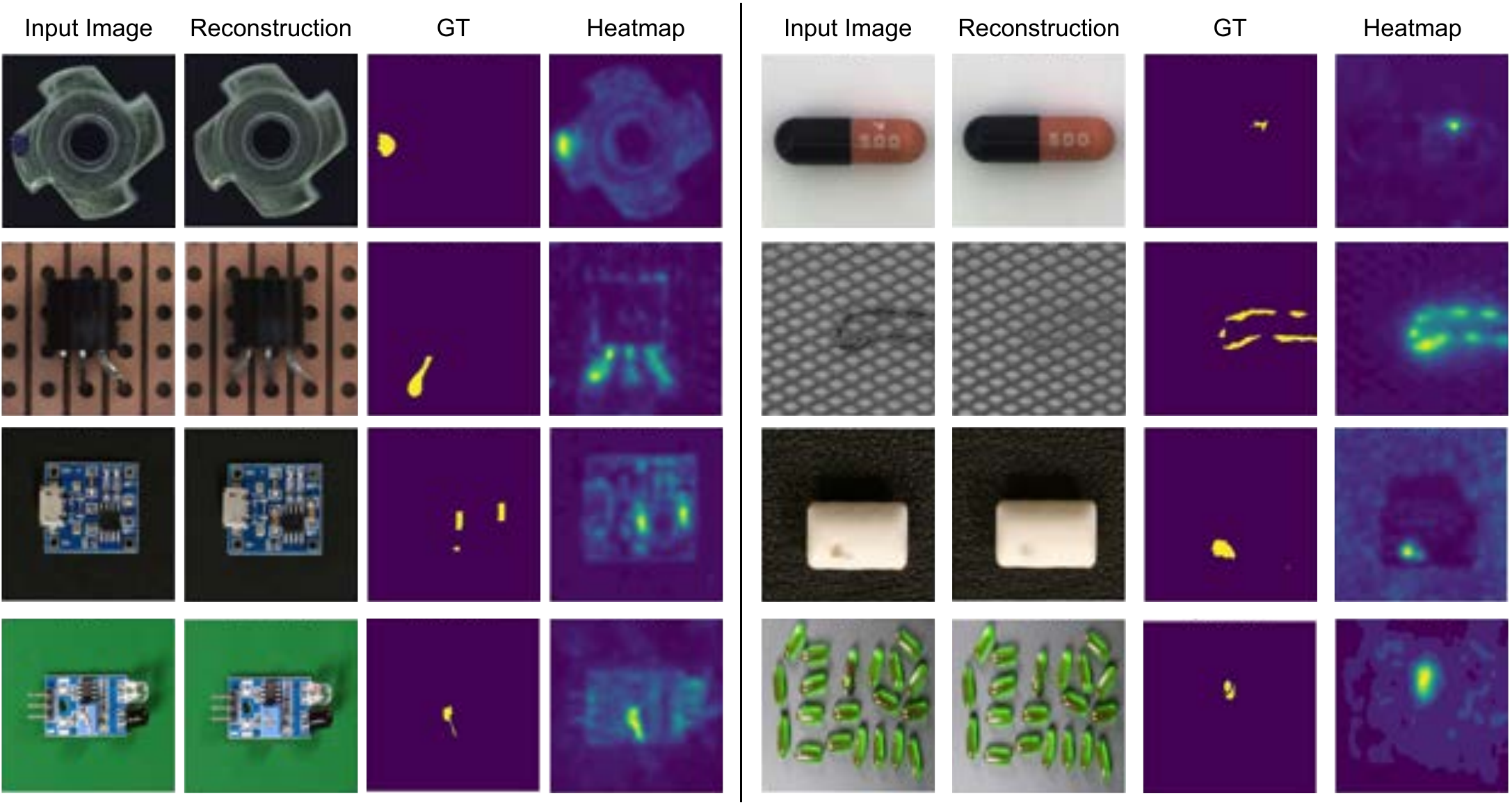}
    \caption{First and second rows depict samples on 'metal nut', 'capsule', 'transistor', and 'grid' selected from MVTec \cite{bergmann2019mvtec}. Third and fourth rows depict samples of 'pcb4', 'chewing gum', 'pcb3' and 'capsules' selected from VisA \cite{zou2022spot}.}
    \label{fig:qualitative}
\end{figure*}

Figures \ref{fig:Teaser} and \ref{fig:qualitative} present the qualitative results obtained for reconstruction and anomaly segmentation. Note that anomalies are localised with remarkable accuracy in various samples of the VisA and MVTec datasets. The model's reconstruction outputs are particularly impressive, as they not only segment anomalous regions but also transform them into their nominal counterparts. For instance, the model regenerates missing links on transistors, erases blemishes on circuit boards, and recreates missing components on PCBs. These reconstructions hold significant value in industrial settings, as they provide valuable insights to workers, enabling them to identify defects and potentially resolve them. Figure \ref{fig:Condition} also qualitatively analyses the impact of conditioning as the hyperparameter \(w\) increases, emphasising that higher values of \(w\) lead to more pronounced conditioning in the reconstructions. Furthermore, this figure also includes a qualitative ablation of the feature-wise and pixel-wise comparisons. More detailed quantitative and qualitative results are included in the Appendix.

\subsection{Inference Time}
The trade-off between accuracy and computation time on the VisA dataset is depicted in Table \ref{DDAD-n}. Among the tested approaches, \textit{DDAD-10} stands out by utilizing 10 iterations and delivering the most favourable results. However, \textit{DDAD-5} becomes an appealing option due to its faster inference time, which holds significant importance, especially in industrial applications. Despite the diffusion model's reputation of slow inference, our approach remains highly competitive with various representation-based models. Our unique conditioning mechanism enables competitive results with fewer denoising steps. This trend holds even with a compressed denoising network (DDAD-S). Our complete DDAD model requires 0.79GB of memory during inference, while DDAD-S only needs 0.59GB including the feature extractor's memory usage.

\input{tables/DDAD-n}

%% file: tables/visaAUROC.tex
\begin{table*}[ht]
\centering
\caption{Anomaly Classification and localisation performance (image AUROC,pixel AUROC) of various methods on VisA benchmark. The best results are highlighted in bold.} 
\resizebox{\textwidth}{!}{
\begin{tabular}{cccccccccccccccc}
\toprule
\textbf{Method} & \textbf{Candle} & \textbf{Capsules} & \textbf{Cashew} & \textbf{Chewing gum} & \textbf{Fryum} & \textbf{Macaroni1} & \textbf{Macaroni2} & \textbf{PCB1} & \textbf{PCB2} & \textbf{PCB3} & \textbf{PCB4} & \textbf{Pipe fryum} & \textbf{Average} \\
\midrule
WinCLIP \cite{Jeong_2023_CVPR} & (95.4,88.9) & (85.0,81.6) & (92.1,84.7) & (96.5,93.3) & (80.3,88.5) & (76.2,70.9) & (63.7,59.3) & (73.6,61.2) & (51.2,71.6) & (73.4,85.3) & (79.6,94.4) & (69.7,75.4) & (78.1,79.6) \\
% SPADE \cite{cohen2020sub} & (91.0,97.9) & (61.4,60.7) & (97.8,86.4) & (85.8,98.6) & (88.6,96.7) & (95.2,96.2) & (87.9,87.5) & (72.1,66.9) & (50.7,71.1) & (90.5,95.1) & (83.1,89.0) & (81.1,81.8) & (82.1,85.6) \\
SPD \cite{zou2022spot}  & (89.1,97.3) & (68.1,86.3) & (90.5,86.1) & (\textbf{99.3},96.9) & (89.8,88.0) & (85.7,98.8) & (70.8,96.0) & (92.7,97.7) & (87.9,97.2) & (85.4,96.7) & (99.1,89.2) & (95.6,95.4) & (87.8,93.8)\\
DRAEM \cite{zavrtanik2021draem} & (91.8,96.6) & (74.7,98.5) & (95.1,83.5) & (94.8,96.8) & (97.4,87.2) & (97.2,\textbf{99.9}) & (85.0,\textbf{99.2}) & (47.6,88.7) & (89.8,91.3) & (92.0,98.0) & (98.6,96.8) & (\textbf{100},98.8) & (88.7,93.5)\\
OmniAL \cite{Zhao_2023_CVPR} & (85.1,90.5) & (87.9,98.6) & (\textbf{97.1},\textbf{98.9}) & (94.9,\textbf{98.7}) & (97.0,89.3) & (96.9,98.9) & (89.9,99.1) & (96.6,\textbf{98.7}) & (99.4,83.2) & (96.9,\textbf{98.4}) & (97.4,\textbf{98.5}) & (91.4,99.1) & (94.2,96.0) \\
DDAD-10 & (\textbf{99.9},\textbf{98.7}) & (\textbf{100},\textbf{99.5}) & (94.5,97.4) & (98.1,96.5) & (\textbf{99.0},\textbf{96.9}) & (\textbf{99.2},98.7) & (\textbf{99.2},98.2) & (\textbf{100},93.4) & (\textbf{99.7},\textbf{97.4}) & (\textbf{97.2},96.3) & (\textbf{100},\textbf{98.5}) & (\textbf{100},\textbf{99.5}) & (\textbf{98.9},\textbf{97.6}) \\
\bottomrule
\end{tabular}
}
\label{Table: VisA AUROC}
\end{table*}

%% file: tables/MTD.tex
\begin{table}[t]
\centering
\caption{Image AUROC results of Anomaly Detection on MTD \cite{huang2020surface}} 
\resizebox{\linewidth}{!}{
\begin{tabular}{cccc}
\toprule
 GANomaly \cite{akcay2019ganomaly} & DifferNet \cite{rudolph2021same} & PatchCore-10 \cite{roth2022towards} & DDAD-10 \\
\midrule
 76.7 & 97.7 & 97.9 & \textbf{98.3} \\
\bottomrule
\end{tabular}}
\label{BTAD MTD AUROC}
\end{table}

%% file: tables/MVTec-VisA-PRO.tex
\begin{table}[!pt]
\centering
\caption{PRO metric for anomaly localisation on MVTec AD \cite{bergmann2019mvtec} and VisA \cite{zou2022spot} dataset. The best results are highlighted in bold.} 
\resizebox{\linewidth}{!}{\begin{tabular}{cccccccccc}
  \toprule
\textbf{Method} & SPADE \cite{cohen2020sub} &PaDiM\cite{defard2021padim} &RD4AD\cite{deng2022anomaly} &PatchCore\cite{roth2022towards} &DDAD-10\\
 \midrule
\textbf{MVTec} & 91.7 & 92.1 & \textbf{93.9} & 93.5& 92.3 \\
 \midrule
 \textbf{Method} & 	
WinCLIP \cite{Jeong_2023_CVPR} &DRAEM \cite{zavrtanik2021draem} &RD4AD\cite{deng2022anomaly} &PatchCore\cite{roth2022towards} &DDAD-10\\
 \midrule
  \textbf{VisA} & 56.8 &73.1  &70.9 & 91.2 & \textbf{92.7} \\
\bottomrule
\end{tabular}}
\label{Table: MVTec-VisA PRO}
\end{table}

%% file: tables/DDAD-n.tex
\begin{table}[ht]
\centering
\caption{Inference time per image and performance of the model on MVTec \cite{bergmann2019mvtec} with different number of denoising steps in the format of (Image AUROC, Pixel AUROC, PRO).} 
\resizebox{\linewidth}{!}{\begin{tabular}{ccccc}
\toprule
\textbf{Method}&PatchCore-1\% &PaDiM  &DDAD-5 &PatchCore-10\% \\
\hline
\textbf{Performance}& (99.0, 98.1, 93.5) &(95.4,97.5,92.1) &(99.3, 97.5, 91.2) &(99.1,98.1,93.5)  \\
\hline
\textbf{Time (s)} &0.17 &0.19 &0.21  &0.22 \\
\toprule

\textbf{Method}  &DDAD-S-10  &DDAD-10  &SPADE  &DDAD-25 \\
\hline
\textbf{Performance}  &(99.7,97.9,91.3)&(99.8,98.1,92.4) &(85.3, 96.6, 91.5) &(99.7, 97.9, 91.0)\\
\hline
\textbf{Time (s)} &0.34  &0.38  &0.66 &0.90 \\
\bottomrule

\end{tabular}}
\label{DDAD-n}
\end{table} 

%% file: sec/6_conclusion.tex
\section{Conclusion}
\label{sec:conclusion}

We have introduced Denoising Diffusion Anomaly Detection (\textit{DDAD}), a new reconstruction-based approach for detecting anomalies. Our model leverages the impressive generative capabilities of recent diffusion models to perform anomaly detection. We design a conditioned denoising process to generate an anomaly-free image that closely resembles the target image. Moreover, we propose an image comparison method based on pixel and feature matching for accurate anomaly localisation. Finally, we introduced a novel technique that utilises our denoising model to adapt a pretrained neural network to the problem's domain for expressive feature extraction. \textit{DDAD} achieves state-of-the-art results on benchmark datasets, namely MVTec, VisA, and MTD, \emph{despite} being a reconstruction-based method.

\textbf{Limitations and future work. } In this work, we demonstrate that our contributions enhance inference speeds while maintaining equivalent anomaly detection performance. Nevertheless, we believe there is still room for improving anomaly localisation. Interventions such as dynamically selecting the denoising starting points or abstracting to a latent space for training are promising avenues to explore in future work.

% In this word, we demonstrated that reducing noise levels and denoising steps enhances inference speed and causes a satisfactory performance. Nevertheless, we noted a potential drawback: this approach could lead to model failure when reconstructing anomalies with significant missing components, resulting in inaccurate anomaly segmentation. Exploring alternatives, such as employing multiple denoising starting points or training the model within the latent space, might offer promising solutions to this challenge. We leave these intriguing avenues for future experimentation.

%% file: sec/X_suppl.tex
\clearpage
\setcounter{page}{1}
\maketitlesupplementary
\section{Implementation Details}
\label{Implementation Details}
DDAD is implemented in Python 3.8 and PyTorch 1.13. The denoising model undergoes training using the Adam optimiser, with a learning rate of 0.0003 and weight decay of 0.05. Fine-tuning of the feature extractor uses an AdamW optimiser with a learning rate of 0.0001. During fine-tuning, each batch is divided into two mini-batches, each of size 16 or 8. One mini-batch consists of input images, while the other comprises target images. The conditioning control parameter is set to \( w=3 \) for fine-tuning the feature extractor. The balance between pixel-wise and feature-wise distance is established as \(v = 1\) for MVTec and \(v = 7\) for VisA. To smooth the anomaly heatmaps, a Gaussian filter with \({\sigma}_g = 4\) is applied. All experiments are executed on a GeForce RTX 3090. The denoising network requires 4 to 6 hours of training, depending on the number of samples for each category.

We obtained the best results using WideResNet101 \cite{zagoruyko2016wide} as the feature extractor. The stochasticity parameter of \(\sigma\) for the denoising process is set equal to 1. Empirically, we achieved similar results in employing a denoising process that is either probabilistic or implicit. Nevertheless, it is essential to note that changing this hyperparameter affects reconstruction, and thus requires additional hyperparameter tuning.

\subsection{MVTec}
In table \ref{Table: replication MVTec} and table \ref{Table: replication MVTec-small}, the settings used to achieve the best result on DDAD and DDAD-S are demonstrated. We have trained DDAD and DDAD-S with a batch size of 32 and 16 respectively. For both models, the feature extracted is fine-tuned and the model is tested on a batch size of 16. Hyperparameter \(v\) is set to 1 to balance pixel and feature comparison. Results on the PRO metric and comparison with the other approaches are depicted in Table \ref{Table:MVTecPRO_Appendix}. Results on different denoising steps are presented in Table \ref{Table: DDAD-n-MVTec}. We have observed setting \({\lambda}_{DL}=0.1\) for the MVTec dataset leads to the best result.

\input{tables/DDAD-n-MVTec}

\subsection{VisA}

Table \ref{Table: replication VisA} showcases the configuration employed to attain optimal results for DDAD. DDAD has undergone training and testing with a batch size of 32. For the categories macaroni2 and pcb1, we achieved better results with a batch size of 16 during fine-tuning. The hyperparameter \(v\) is established at 7; however, setting \(v\) to 1.5 for cashew yields a more precise detection. Results on the PRO metric are depicted in Table \ref{Table:VisAPRO_Appendix}. We have observed setting \({\lambda}_{DL}=0.01\) for the VisA dataset leads to the best result.

\input{tables/PRO}

\input{tables/replication}

\section{Hyperparameters}
\label{Hyperparameters}
In this section, we discuss the role of each hyperparameter introduced in the paper and how they solely affect the quality of reconstruction or precision of the localisation heatmap.

\subsection{Conditioning hyperparameter w}
\label{Rule of the conditioning hyperparameter w}
\begin{figure*}[ht]
    \centering
    \includegraphics[width = \textwidth]{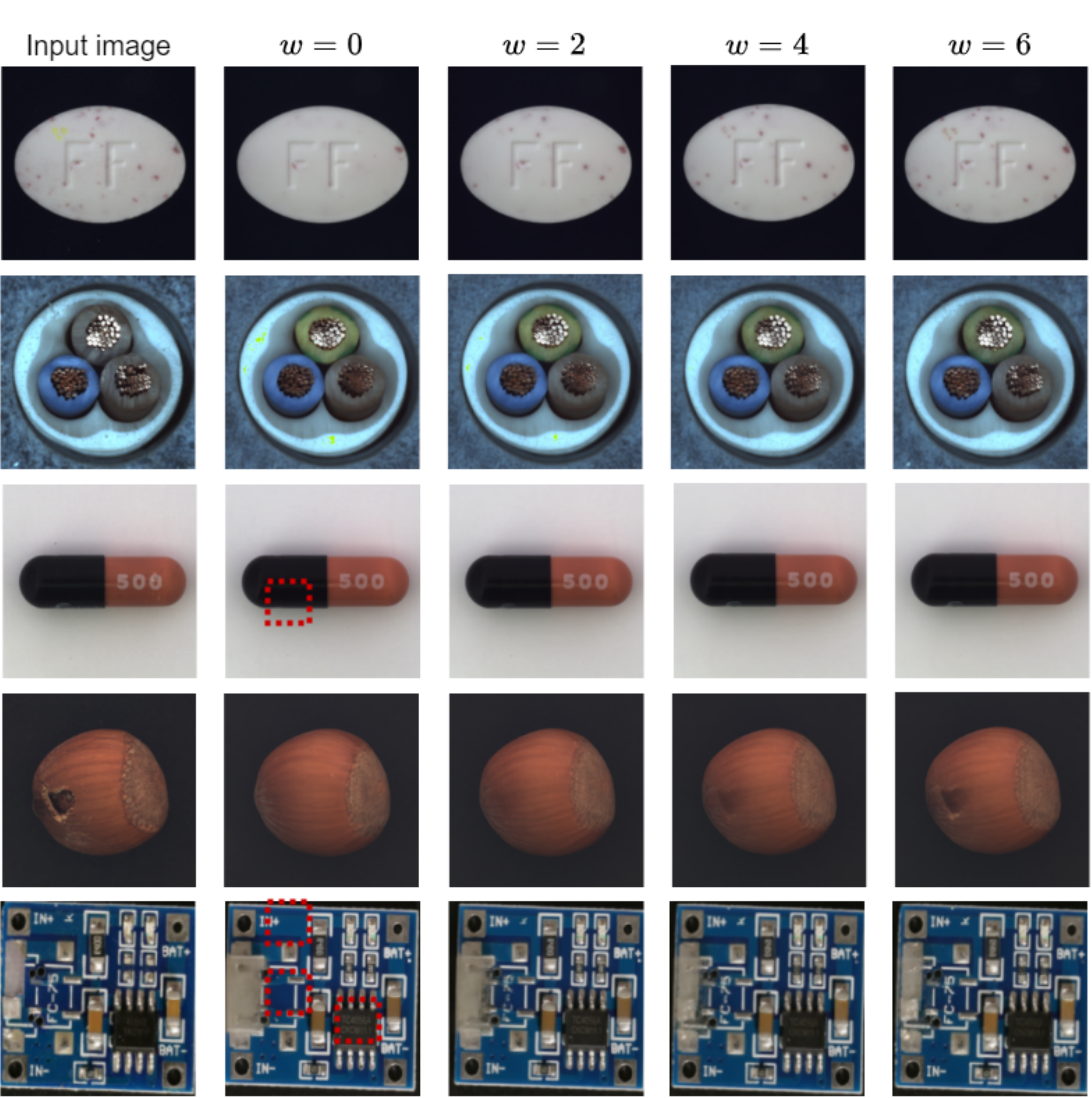}
    \caption{Some qualitative results, showcasing the insufficiency of plain diffusion models for more accurate anomaly detection.}
    \label{fig:conditioning_qualitative_appendix}
\end{figure*}

Table \ref{Table:w_parameter} presents quantitative results on the impact of the hyperparameter \(w\) on enhanced reconstruction, illustrating how the conditioning mechanism reduces misclassification and mislocalisation across 13 out of 15 categories. To ensure a fair comparison, we exclusively used pixel-wise distance to assess the reconstruction quality on the MVTec dataset. As shown in Figure \ref{fig:Comparison_methods_MVTec} (left), this conditioning improves anomaly detection and localisation by 6.7\% and 4.2\%, respectively. Notably, in some categories, such as pill and tile, our conditioning mechanism enhances reconstruction by up to 30\%. The same improvement is observed in Figure \ref{fig:BarChart_VisA_Appendix} when our conditioning mechanism is applied in the denoising process. Figure \ref{fig:conditioning_qualitative_appendix} qualitatively illustrates the impact of conditioning on reconstruction.

By introducing the conditioning mechanism, we achieve reconstruction of anomalous regions while effectively preserving the pattern of nominal regions. In the provided example, the first row displays a sample from the \textit{pill} category of the MVTec dataset \cite{bergmann2019mvtec}, where red dots are often randomly distributed. A plain diffusion model fails to accurately reconstruct the dots. However, by increasing the conditioning parameter \(w\), the model successfully reconstructs these red dots, simultaneously eliminating and replacing the anomaly (yellow colour on the top left side of the pill) with the nominal pattern.

In the second row, an example of a cable is shown, where the plain diffusion correctly changed the colour of the top grey cable to green. However, compared to the conditioned reconstruction, where the wires are accurately reconstructed, the plain diffusion model failed to correctly reconstruct the individual wires within the cable. In the third row, there is an example of a printed part, indicated by a red box, on the capsule that is not successfully reconstructed using a plain diffusion model. However, when conditioning is applied, the printed part is restored to its original form.

In the case of the \textit{hazelnut}, the plain diffusion model results in a rotated reconstruction, which is incorrect. When conditioning is applied, the rotation is effectively corrected, and the hazelnut is reconstructed in the right orientation. Additionally, the rays on the hazelnut are reconstructed similarly to the input image, maintaining their original appearance. The last row showcases an example from the VisA dataset \cite{zou2022spot}. After the reconstruction process, certain normal parts highlighted by the red boxes are eliminated. This absence of information is rectified through the conditioning of the model on the input image, allowing the model to accurately reconstruct these areas. The conditioning mechanism plays a crucial role in preventing these alterations from being erroneously flagged as anomalous patterns, ensuring precision in the reconstruction process.

\input{tables/pixel-based-comparison}

\subsection{Hyperparameter v}
In two tables, \ref{Table: vparameter-mvtec} and \ref{Table: vparameter-visa}, we elucidate the influence of the hyperparameter \(v\) on the amalgamation of pixel-wise and feature-wise comparisons. Most categories demonstrate that minor adjustments to this hyperparameter do not yield significant changes. This observation suggests that the combination technique accommodates a broad spectrum of anomalies and is not highly sensitive to the \(v\) hyperparameter. Nevertheless, we fine-tuned this hyperparameter to optimise results.

\input{tables/v-parameter}

\section{Ablation on VisA}
\label{Ablation on VisA}
\begin{figure*}[ht]
\centering
\includegraphics[width=\textwidth]{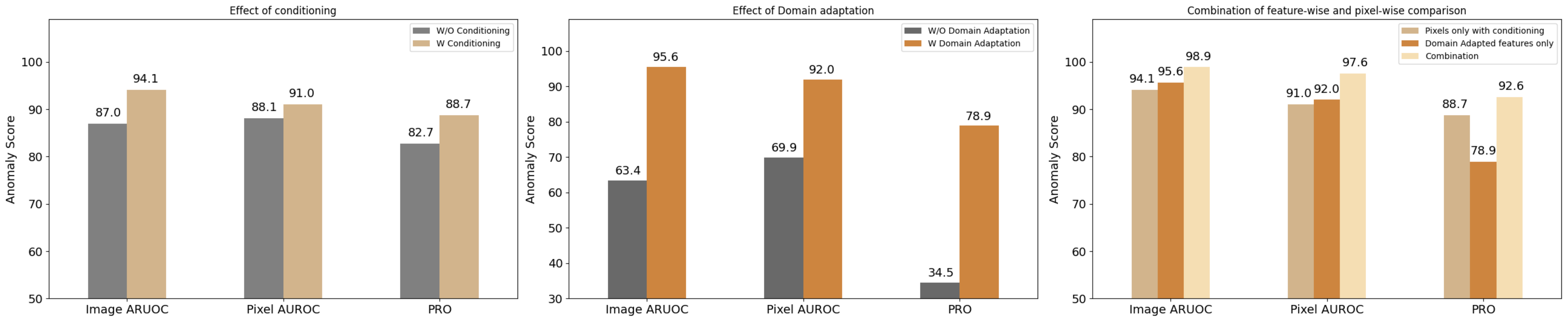}
\caption{Effectiveness of various components of our model on anomaly detection and segmentation. Left: Effectiveness of conditioning based on pixel-wise image comparison. Middle: Performance increase due to domain adaptation of feature extractor. Right: Impact of merging feature-wise and pixel-wise image comparison. All results are shown on the VisA \cite{zou2022spot} dataset.}
\label{fig:BarChart_VisA_Appendix}
\end{figure*}

As demonstrated in Section \ref{Experimental Results and Discussions}, our conditioning approach significantly enhances the model's performance compared to plain diffusion models. This improvement on MVTec \cite{bergmann2019mvtec} is also evident in Figure \ref{fig:BarChart_VisA_Appendix}, where the image AUROC, pixel AUROC, and PRO metrics have increased by \(7.1\%\), \(2.9\%\), and \(6.0\%\), respectively, using pixel-wise comparison. While pixel-wise comparison alone achieves promising results of \(94.1\%\), the overall performance increases to \(98.9\%\) after combining it with feature-wise comparison. We observed that a pretrained feature extractor performs poorly in feature-wise comparison. However, these results have significantly improved after domain adaptation. Image AUROC, pixel AUROC, and PRO metrics increased by \(32.2\%\), \(22.1\%\), and \(44.4\%\), respectively, when only feature-wise comparison is used. The inability of the pretrained feature extractor to extract informative features may explain the inferior performance of representation-based models compared to DDAD, where the backbone fails to provide better features. Detailed performance of feature-wise and pixel-wise comparison for each category are shown in tables \ref{Table: W-WO conditioning VisA} and \ref{Table: W-WO domain adaptatoin VisA}, respectively.

\input{tables/W-WO-VisA}

\section{Feature Extractor}
\input{tables/feature_extractor}
\subsection{Different backbones}
Tables \ref{Table:feature-extractor-MVTec} and \ref{Table:feature-extractor-visA} provide a detailed analysis of results obtained using various backbones as the feature extractor. Notably, while WideResNet101 yielded the best outcomes for both MVTec and VisA, WideResNet50 demonstrated comparable results.

\subsection{Distillation loss to not forget}
\begin{figure*}[ht]
\centering
\includegraphics[width=\textwidth]{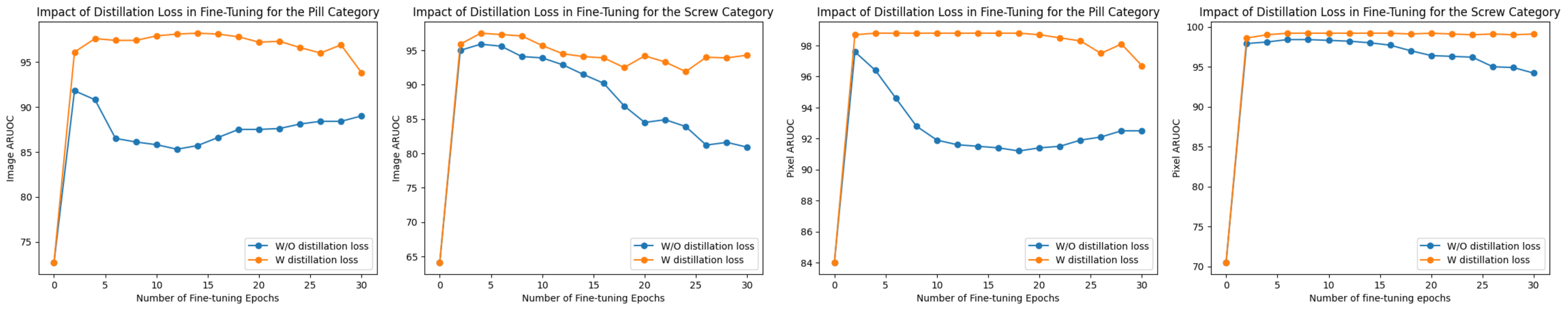}
\caption{Role of distillation loss in fine-tuning to avoid pretrained knowledge loss.}
\label{fig:W-WO-Distillation}
\end{figure*}

As quantitatively illustrated in Table \ref{Table: lambda-DL-MVTec}, undertaking domain adaptation without incorporating a distillation loss leads to the feature extractor erasing its prior knowledge. It is crucial to retain the pretrained information during the transition to a new domain, as the feature extractor's capability to discern anomalous features is rooted in its training on extensive data on ImageNet. We exemplify this phenomenon with Pill and Screw categories in Figure \ref{fig:W-WO-Distillation}. The figure showcases how the introduction of distillation loss prevents AUROC deterioration over epochs, indicating that the feature extractor adapts to the new domain while preserving its pretrained knowledge. In the absence of distillation loss, the feature extractor begins to lose its generality, a critical aspect for extracting anomalous features.

\subsection{Robustness to anomalies on the background}
\label{Robustness to anomalies on the background}
\begin{figure*}[ht]
    \centering
    \includegraphics[width = \textwidth]{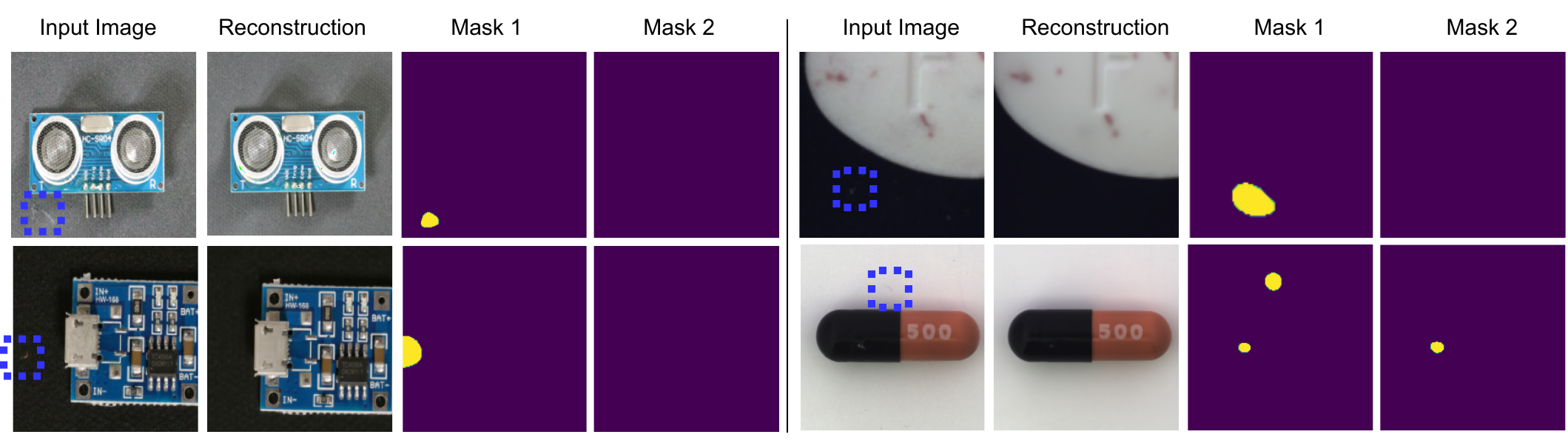}
    \caption{Some qualitative results where the background patterns are considered anomalies when a pretrained feature extractor is used. It is shown by Mask 1 in Figure. After domain adaptation, the feature extractor becomes robust to these changes. It is shown by Mask 2.}
    \label{fig:FE_effect_appendix}
\end{figure*}

In industrial and production scenarios, a significant challenge often involves dealing with anomalies, such as dust or environmental changes in the background during photography. In this section, we highlight the robustness of the domain-adapted feature extractor to such spurious patterns. As depicted in Figure \ref{fig:FE_effect_appendix}, a pretrained feature extractor erroneously identifies normal background elements, indicated by the blue boxes, as anomalies. However, after domain adaptation, the feature extractor becomes resilient, no longer misidentifying or mislocating these elements. In the first three samples, showcasing PCBs, not only are anomalies mislocalised, but the images are also misclassified.

\input{tables/parameter-lambda}

\section{Comparative Analysis of Present Diffusion-Based Anomaly Detection Models}
In this section, we compare our model with similar approaches that utilise denoising diffusion models for anomaly detection. We showcase a unique aspect of our architecture that sets it apart from others and demonstrates superior performance.

AnoDDPM \cite{wyatt2022anoddpm} demonstrated that starting from a full-length Markovian chain is not imperative. Additionally, they demonstrated that a multi-scale simplex noise leads to a better reconstruction.

However, substituting Gaussian noise with simplex noise results in a slower inference time. Generally, the time complexity of sampling simplex noise, which is \(\mathcal{O}(n^2)\), is typically higher than that of Gaussian noise, which is \(\mathcal{O}(1)\), due to its inherent complexity. While the time complexity of simplex noise is often discussed in terms of operations per sample, varying with implementation details and dimensions, Gaussian noise generation using conventional methods is commonly considered constant time, \(\mathcal{O}(1)\), per sample. To avoid this replacement, we introduced a conditioning mechanism that enables us to initiate from higher time steps. This allows for the reconstruction of components situated in low distribution while preserving the nominal part of the image.

DiffusionAD \cite{zhang2023diffusionad}, developed concurrently with this work, employs two sub-networks for denoising and segmentation, inspired by DRAEM \cite{zavrtanik2021draem}, showcasing the success of diffusion models over VAEs in anomaly detection. While a single denoising step accelerates the process, it makes it akin to VAEs, moving directly from noise to signal, with the distinction that in this case, the starting point is a noise-to-signal ratio. Additionally, they rely on external synthetic anomalies, potentially decreasing robustness to unseen anomalies. According to the results, DDAD outperforms by 1.1\% on the Image AUROC metric for the VisA dataset. Results on pixel AUROC are not published.

Score-based perturbation resilience \cite{Shin_2023_ICCV} formulates the problem with a geometric perspective. The idea is based on the assumption that samples that deviate from the manifold of normal data, cannot be restored in the same way as normal samples. Hence, the gradient of the log-likelihood results in identifying anomalies. Score-based perturbation resilience, unlike DiffusionAD and DRAEM, does not rely on any external data, making them robust to a wide range of anomalies. However, this approach fails to outperform representation-based models in both anomaly segmentation and localisation. According to the results, DDAD outperforms by 2.1\% and 0.7\% on the Image AUROC and Pixel AUROC metrics for the MVTec dataset.

Lu et al. \cite{Lu_2023_ICCV} leverage the KL divergence between the posterior distribution and estimated distribution as the pixel-level anomaly score. Additionally, an MSE error for feature reconstruction serves as a feature-level score. This model relies on a pretrained feature extractor, which may not be adapted to the domain of the problem. Moreover, the outcomes are not competitive with representation-based models. DDAD outperforms by 1.4\% on the Pixel AUROC metric for the MVTec dataset. Results on Image AUROC are not published.

To avoid reliance on external resources, we introduce a domain adaptation technique to address the domain shift problem. Additionally, a guidance mechanism is introduced to tailor the denoising process for the task of anomaly detection, preserving the nominal part of the image. Notably, the aforementioned papers did not benchmark on both MVTec and VisA, nor were they evaluated based on all three metrics: Image AUROC, Pixel AUROC, and PRO. In this paper, we demonstrate the robustness of our model through a comprehensive analysis of both MVTec and VisA, evaluating all three metrics. We show that DDAD not only outperforms reconstruction-based models but also representation-based models. 

\section{The Importance of Combining Pixel-wise and Feature-wise Comparison}
\label{Need for both pixel and feature comparison.}
\begin{figure*}[ht]
    \centering
    \includegraphics[width=\textwidth]{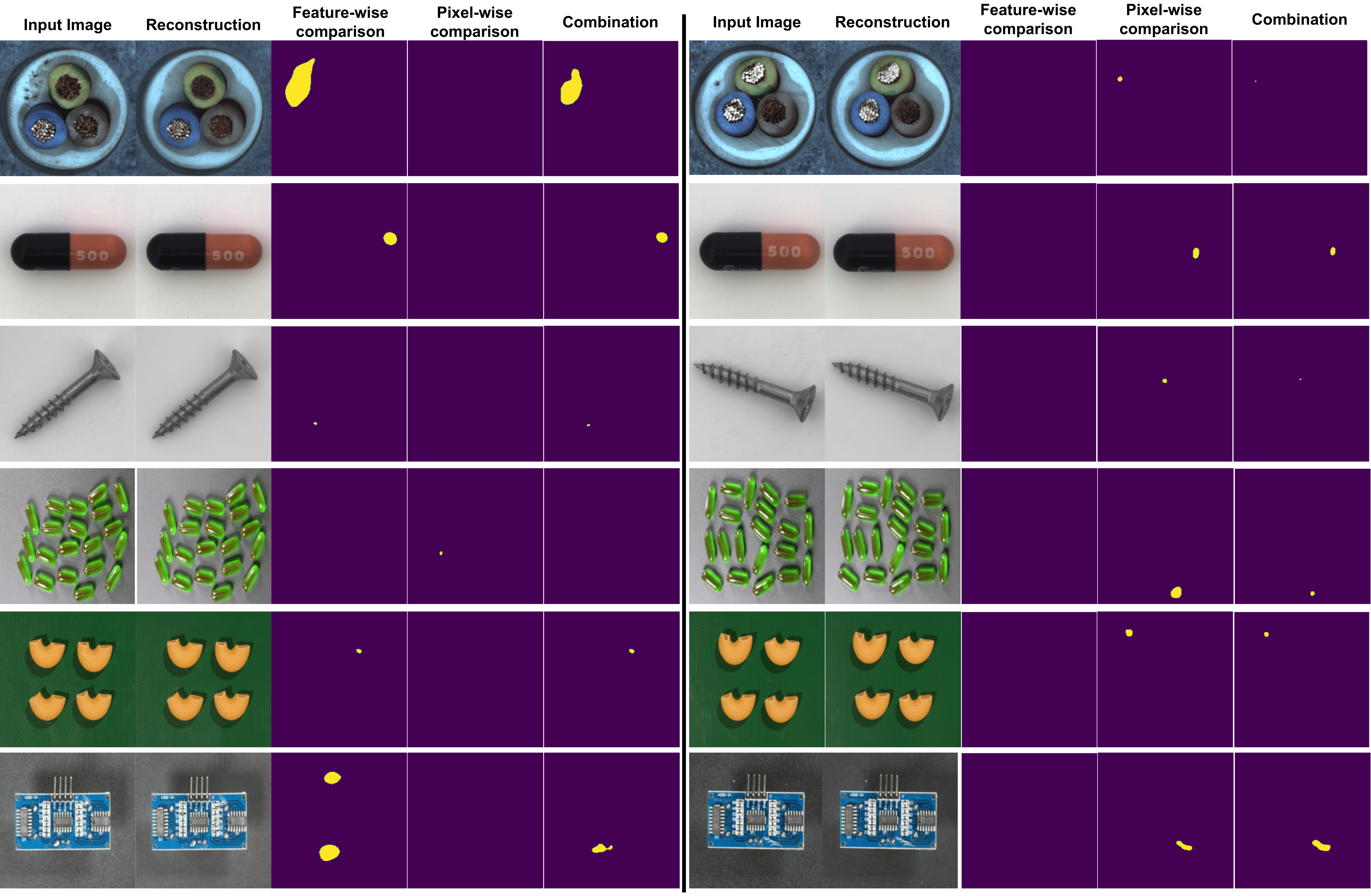}
    \caption{The left-side examples depict cases where pixel-wise comparison fails to accurately detect and pinpoint anomalies, while a feature-wise comparison successfully highlights these anomalies. In contrast, the right-side examples demonstrate situations where feature-wise comparison falls short, yet pixel-wise comparison excels in detecting anomalies.}
    \label{fig: combination of feature and pixel}
\end{figure*}

In Figure \ref{fig: combination of feature and pixel}, we present six examples from MVTec (top three rows) and six examples from VisA (last three rows), where either pixel-wise or feature-wise comparison proves ineffective. In the last row, the initial PCB example fails in both scenarios. The feature-wise comparison identifies two anomalous regions, whereas the pixel-wise comparison does not identify any region as an anomaly. Intriguingly, after combining the approaches, the score of the region previously misidentified as an anomaly decreases. This region is now segmented as normal after the combination.

\section{Additional Qualitative Results}

\subsection{Mislocalisation} 
\label{Mislocalisation}
\begin{figure*}[ht]
    \centering
    \includegraphics[width = \textwidth]{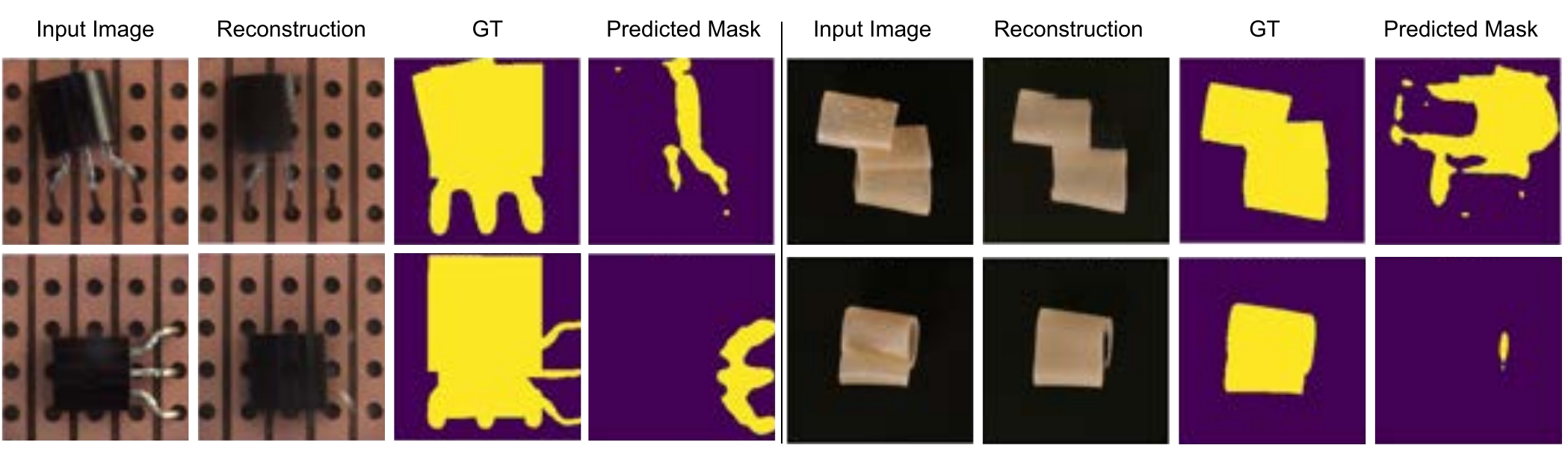}
    \caption{Transistor samples from MVTec \cite{bergmann2019mvtec} and pipe fryum examples from \cite{zou2022spot}. While the images are correctly classified, there is a huge mislocalisation.}
    \label{figure:Failures_appendix}
\end{figure*}

Although our model achieved a high AUROC for anomaly detection, it faced challenges in accurately localising extreme rotations or figure alterations. For example, as depicted in Figure \ref{figure:Failures_appendix}, when starting from time step 250, the model struggled to reconstruct these substantial changes. Conversely, beginning from larger time steps make the reconstruction process difficult and slow. Additionally, it is important to note that our conditioning approach aims to preserve the overall structure of the reconstructed image similar to the input image. However, in cases where there are drastic changes such as rotations or figure alterations, the conditioning mechanism may lead to mislocalisation.

\subsection{Qualitative results on MTD}

\begin{figure*}[ht]
    \centering
    \includegraphics[width = \textwidth]{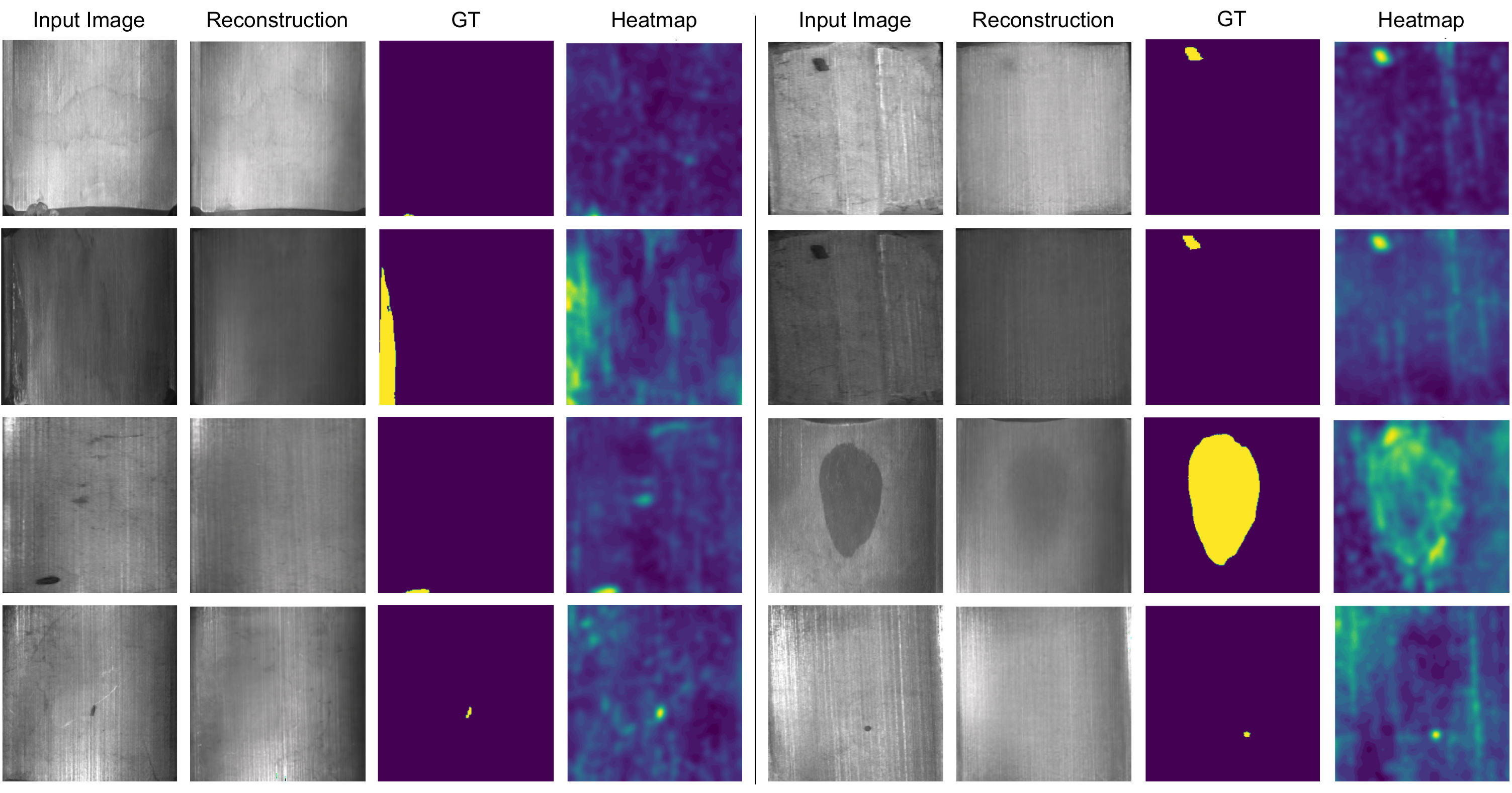}
    \caption{MTD dataset \cite{huang2020surface}.}
    \label{figure:MTD_Appendix}
\end{figure*}

To showcase the versatility of our model beyond the MVTec \cite{bergmann2019mvtec} and VisA \cite{zou2022spot} datasets, we also evaluated DDAD performance on an entirely different dataset called MTD \cite{huang2020surface}. This evaluation allows us to demonstrate the potential of our model across diverse datasets. In Figure \ref{figure:MTD_Appendix}, we present qualitative results illustrating the performance of our DDAD approach on the MTD dataset. 

%% file: tables/DDAD-n-MVTec.tex
\begin{table*}[ht]
\centering
\caption{DDAD Performance on MVTec \cite{bergmann2019mvtec}, based on various denoising steps.Format (ImageAUROC, PixelAUROC)} 
\resizebox{\textwidth}{!}{\begin{tabular}{ccccccccccccccccc}
\toprule
 \textbf{Categories} &Carpet &Grid &Leather & Tile &Wood &Bottle &Cable &Capsule &Hazelnut & Metal nut &Pill &Screw &Toothbrush &Transistor &Zipper &Avg\\
 \midrule
 DDAD-5  &(94.3,96.4) &(100,99.3) &(100,99.1) & (100,98.2) &(99.5,94.4) &(100,98.7) &(99.6,98.2) &(99.1,93.8) &(100,98.2) & (99.7,98.0) &(99.9,98.8) &(97.4,98.9) & (100,98.6) &(99.8,94.0) &(100,98.3) &(99.3, 97.5) \\
 DDAD-10 &(99.3,98.7) & (100,99.4) &(100,99.4) & (100,98.2) &(100,95.3) &(100,98.7) &(99.4,98.1) &(99.4,95.7) &(100,98.3) &(100,98.9) &(100,99.1) &(99.0,99.3) &(100,98.7) &(100,95.3) & (100,98.2) &(99.8,98.1) \\
 DDAD-25   &(99.0,98.7) &(100,99.3) &(100,99.0) &  (100,98.3) &(99.4,94.2) &(100,98.7) &(99.6,98.2) &(99.6,95.4) &(99.9,98.2) & (99.5,98.7) &(100,98.9) &(99.1,99.3)&(100,98.7) & (100,95.0) &(100,98.2) &(99.7, 97.9) \\
DDAD-S-10 &(98.2,98.6) &(100,98.4) &(100,99.2) & (100,98.2) &(99.9,95.1) & (100,98.5) &(99.8,98.3) &(99.4,96.0) &(99.8,98.4) & (100,98.1) &(99.5,99.1) &(98.3,99.0) &(100,98.7) &(100,95.3) &(99.9,97.5) &(99.7,97.9)\\
\bottomrule
\end{tabular}}
\label{Table: DDAD-n-MVTec}
\end{table*}

%% file: tables/PRO.tex
\begin{table*}[ht]
\centering
\caption{Anomaly Localisation Performance on MVTec \cite{bergmann2019mvtec}, based on PRO metric.} 
\resizebox{\textwidth}{!}{\begin{tabular}{ccccccccccccccccc}
\toprule
\textbf{Categories} &Carpet &Grid &Leather & Tile &Wood &Bottle &Cable &Capsule &Hazelnut & Metal nut &Pill &Screw &Toothbrush &Transistor &Zipper &Avg\\
\midrule
\textbf{SPADE \cite{cohen2020sub}} &94.7 &86.7 &97.2 & 75.9 &87.4 &95.5 &90.9 &93.7 &95.4  &\textbf{94.4} &94.6 &96.0 &93.5 &\textbf{87.4} &92.6 &91.7 \\
\textbf{PaDiM \cite{defard2021padim}} &96.2 &94.6 &97.8 & 86.0 &\textbf{91.1} &94.8 &88.8 &93.5 &92.6 & 85.6 &92.7 &94.4 &93.1 &84.5 &95.9 &92.1 \\
\textbf{RD4AD \cite{deng2022anomaly}} &\textbf{97.0} &97.6 &\textbf{99.1} & 90.6 &90.9 &\textbf{96.6} &91.0 &\textbf{95.8} &\textbf{95.5} & 92.3 &\textbf{96.4} &\textbf{98.2} &94.5 &78.0 &95.4 &\textbf{93.9} \\
\textbf{PatchCore \cite{roth2022towards}} &96.6 &95.9 &98.9 & 87.4 &89.6 &96.1 &\textbf{92.6} &95.5 &93.9 & 91.3 &94.1 &97.9 &91.4 &83.5 &\textbf{97.1} &93.5 \\
\midrule
 \textbf{DDAD-5} &86.8 &96.4 &97.2 & 93.1 &82.1 &91.8 & 90.2 &92.5 &87.5 & 88.1 &94.3 &94.7 &91.8 &87.3 &93.9 &91.2 \\
\textbf{DDAD-10} &93.9 &97.3 &97.7 & 93.1 &82.9 &91.8 &88.9 &93.4 &86.7 & 91.1 &95.5 &96.3 &92.6 &90.1 &93.2 &92.3\\
\textbf{DDAD-25} &94.2 &97.0 &97.9 & 84.1 & 77.5 &92.3 &87.4 &91.0 &86.0 & 91.6 &94.9 &95.9 &92.9 &90.4 &92.4 &91.0\\
\textbf{DDAD-S-10} &93.7 &93.9 &96.5 & 93.2 &84.3 &90.6 &87.6 &91.6 &85.4 & 87.4 &95.1 &96.9 &92.4 &91.8 &88.6 &91.3\\
\bottomrule
\end{tabular}}
\label{Table:MVTecPRO_Appendix}
\end{table*}

\begin{table*}[ht]
\centering
\caption{Anomaly Localization Performance on VisA \cite{zou2022spot}, based on PRO metric.} 
\resizebox{\textwidth}{!}{\begin{tabular}{ccccccccccccccc}
\toprule
 \textbf{Categories} &Candle & Capsules &Cashew &Chewing gum &Fryum &Macaroni1 &Macaroni2 &PCB1 &PCB2 &PCB3 &PCB4 &Pipe fryum &Avg\\
 \midrule
 \textbf{SPADE \cite{cohen2020sub}}  &93.2 & 36.1 &57.4 &\textbf{93.9} &91.3 &61.3 &63.4 &38.4 &42.2 &80.3 &71.6 &61.7 &65.9 \\
 \textbf{PaDiM \cite{defard2021padim}}  &95.7 & 74.9 &87.9 &83.5 &80.2 &92.1 &75.4 &91.3 &88.7 &84.9 &81.6 &92.5 &85.9 \\
 \textbf{RD4AD \cite{deng2022anomaly}} &92.2 & 56.9 &79.0 &92.5 &81.0 &71.9 &68.0 &43.2 &46.4 &80.3 &72.2 &68.3 &70.9 \\
 \textbf{PatchCore \cite{roth2022towards}} &94.0 & 85.5 &\textbf{94.5} &84.6 &\textbf{95.3} &95.4 &94.4 &\textbf{94.3} &89.2 &\textbf{90.9} &90.1 &\textbf{95.7} &91.2 \\
  \midrule
 \textbf{DDAD-10} &\textbf{96.6} & \textbf{95.0} & 80.3 &85.2 &94.2 &\textbf{98.5} &\textbf{99.3} &93.3 &\textbf{93.3} &86.6 &\textbf{95.5} &94.7 &\textbf{92.7}\\
\bottomrule
\end{tabular}}
\label{Table:VisAPRO_Appendix}
\end{table*}

%% file: tables/replication.tex
\begin{table*}[ht]
\centering
\caption{Setting for replicating results on MVTec \cite{bergmann2019mvtec}.} 
\resizebox{\textwidth}{!}{\begin{tabular}{cccccccccccccccc}
\toprule
 \textbf{Categories} &Carpet &Grid &Leather & Tile &Wood &Bottle &Cable &Capsule &Hazelnut & Metal nut &Pill &Screw &Toothbrush &Transistor &Zipper\\
 
  \midrule
\(w\)   &0 &4 &11 & 4 &11 &3 &3 &8 &5 & 7 &9 &2 &0 &0 &10\\

Training epochs  &2500 &2000 &2000 & 1000 &2000 &1000 &3000 &1500 &2000 & 3000 &1000 &2000 &2000 &2000 &1000\\

FE epochs  &0 &6 &8 & 0 &16 &5 &0 &8 &3 & 1 &4 &4 &2 &0 &6\\
\bottomrule
\end{tabular}}
\label{Table: replication MVTec}
\end{table*}

\begin{table*}[ht]
\centering
\caption{Setting for replicating results on MVTec \cite{bergmann2019mvtec} for the small model.} 
\resizebox{\textwidth}{!}{\begin{tabular}{cccccccccccccccc}
\toprule
 \textbf{Categories} &Carpet &Grid &Leather & Tile &Wood &Bottle &Cable &Capsule &Hazelnut & Metal nut &Pill &Screw &Toothbrush &Transistor &Zipper\\
 
  \midrule
\(w\)  &0 &5 &6 & 4 &4 &8 &0 &11 &0 & 3 &11 &2 &1 &1 &5 \\

Training epochs &2000 &2000 &2000 & 2000 &2000 &2000 &4000 &3000 &2000 & 2000 &1000 &2000 &2000 &4000 &2000 \\

FE epochs   &0 &4 &4 & 0 &11 &1 &0 &4 &2 & 3 &6 &- &2 &7 &4\\
\bottomrule
\end{tabular}}
\label{Table: replication MVTec-small}
\end{table*}

\begin{table*}[ht]
\centering
\caption{Setting for replicating results on VisA \cite{zou2022spot}.} 
\resizebox{\textwidth}{!}{\begin{tabular}{ccccccccccccc}
\toprule
 \textbf{Categories} &Candle & Capsules &Cashew &Chewing gum &Fryum &Macaroni1 &Macaroni2 &PCB1 &PCB2 &PCB3 &PCB4 &Pipe fryum \\
 
  \midrule
\(w\)  &6 & 5 &0 &6 &4 &5 &2 &9 &5 &6 &6 &8\\

Training epochs  &1000 & 1000 &1750 &1250 &1000 &500 &500 &500 &500 &500 &500 &500  \\

FE epochs  &1 & 3 &0 &0 &3 &7 &11 &8 &5 &1 &1 &6 \\
\bottomrule
\end{tabular}}
\label{Table: replication VisA}
\end{table*} 

%% file: tables/pixel-based-comparison.tex
\begin{table*}[ht]
\centering
\caption{Sensitivity of conditioning parameter \(w\) on MVTec \cite{bergmann2019mvtec} only when compared in pixel-wise distance. Format (ImageAUROC, PixelAUROC)} 
\resizebox{\textwidth}{!}{\begin{tabular}{cccccccccccccccc}
\toprule
 \textbf{Categories} &Carpet &Grid &Leather & Tile &Wood &Bottle &Cable &Capsule &Hazelnut & Metal nut &Pill &Screw &Toothbrush &Transistor &Zipper\\
 \midrule
\(w=0\)  &(66.7,82.6) &(100,99.2) &(99.9,98.9) & (66.6,64.8) &(93.6,81.9) &(96.3,87.5) &(61.2,89.0) &(80.7,76.9) &(95.0,95.5) & (79.1,90.7) &(69.5,80.9) &(96.5,98.8) &(99.7,97.6) &(82.1,82.5) &(99.2,96.3)\\
\(w=1\)  &(69.5,83.6) &(100,99.4) &(99.9,99.1) & (75.6,72.1) &(94.4,83.5) &(96.3,89.8) &(63.3,87.9) &(84.8,85.5) &(96.5,96.8) & (82.9,90.9) &(76.5,89.6) &(97.7,99.1) &(100,97.8) &(85.9,84.0) &(99.7,97.1)\\
\(w=2\)  &(73.4,84.9) &(100,99.5) &(100,99.2) & (86.0,78.8) &(96.7,84.8) &(96.0,90.9) &(69.4,86.5) &(86.8,90.3) &(97.1,97.2) & (85.0,90.3) &(85.0,94.1) &(98.6,99.2) &(99.7,97.9) &(87.0,84.9) &(99.8,97.7)\\
\(w=3\) &(77.0,85.5) &(100,99.6) &(100,99.2) & (92.9,83.9) &(96.8,85.8) &(95.2,91.2) &(73.5,85.1) &(89.7,92.4) &(97.2,97.4) & (86.0,89.2) &(90.2,95.7) &(99.1,99.3) &(99.2,97.9) &(86.1,85.0) &(99.9,98.0)\\
\(w=4\) &(79.3,86.2) &(100,99.6) &(100,99.3) & (96.3,87.3) &(96.8,86.7) &(94.2,91.0) &(74.4,83.8) &(91.1,92.9) &(97.6,97.5) &(86.9,87.9)&(92.6,96.6) &(99.2,99.3) &(98.9,97.8) &(85.7,84.8) &(99.9,98.2)\\
\(w=5\) & (79.4,86.7) &(100,99.6) &(100,99.3) & (98.3,89.4) &(97.1,87.4) &(93.0,90.7) &(75.4,82.8) &(92.2,92.9) &(97.6,97.6) & (87.7,86.5) &(94.1,97.1) &(99.4,99.3) &(97.8,97.7) &(85.2,84.5) &(99.9,98.4)\\
\(w=6\) &(79.9,87.1) & (100,99.6) &(100,99.4) & (98.5,90.7) & (97.3,88.0) &(92.6,90.3) &(77.0,81.9) &(93.1,92.7) &(97.7,97.6) &(87.6,85.3) &(94.3,97.5) & (99.5,99.3) &(96.7,97.6) &(83.6,84.2) &(100,98.5)\\
\bottomrule
\end{tabular}}
\label{Table:w_parameter}
\end{table*} 

%% file: tables/v-parameter.tex
\begin{table*}[ht]
\centering
\caption{Detailed results on parameter \(v\) on MVTec \cite{bergmann2019mvtec}. The format for AUROC is (Image AUROC, Pixel AUROC)} 
\resizebox{\textwidth}{!}{\begin{tabular}{ccccccccccccccccc}
\toprule
 \textbf{Categories} &Carpet &Grid &Leather & Tile &Wood &Bottle &Cable &Capsule &Hazelnut & Metal nut &Pill &Screw &Toothbrush &Transistor &Zipper &Avg\\
 \midrule
 \(v=0.8\) &(99.4,98.8) &(100,99.3) &(100,99.4) & (100,98.2) &(99.7,95.0) &(100,98.7) &(99.3,98.1) &(99.4,95.7) &(100,98.1) & (99.9,98.9) &(100,99.1) &(98.8,99.3) &(100,98.6) &(92.6,91.5) &(100,98.3) &(99.3,97.8)\\
\(v = 1.0\)   &(99.3,98.7) & (100,99.4) &(100,99.4) & (100,98.2) &(100,95.0) &(100,98.7) &(99.4,98.1) &(99.4,95.7) &(100,98.3) &(100,98.9) &(100,99.1) &(99.0,99.3) &(100,98.7) &(100,95.3) & (100,98.2) &(99.8,98.1) \\
\(v=2.0\) &(98.4,98.4) &(100,99.4) &(100,99.4) & (100,98.3) &(100,94.6) &(100,98.7) &(98.8,98.0) &(98.9,95.9) &(99.9,98.7) & (98.4,98.8) &(100,98.8) &(99.2,99.4) &(100,98.8) &(98.7,92.0) &(100,97.6)& (99.5, 97.8)\\
\bottomrule
\end{tabular}}
\label{Table: vparameter-mvtec}
\end{table*}

\begin{table*}[ht]
\centering
\caption{Detailed results on parameter \(v\) on VisA \cite{zou2022spot}. The format for AUROC is (Image AUROC, Pixel AUROC)} 
\resizebox{\textwidth}{!}{\begin{tabular}{cccccccccccccc}
\toprule
\textbf{Categories} &Candle & Capsules &Cashew &Chewing gum &Fryum &Macaroni1 &Macaroni2 &PCB1 &PCB2 &PCB3 &PCB4 &Pipe fryum \\
 \midrule
v=5.0 (AUROC)  &(99.8,98.7) & (100.0,99.4) &(98.3,96.8) &(98.3,96.8) & (99.0,96.7) &(99.3,98.8) &(99.1,98.5) &(99.9,94.1) &(99.8,97.0) &(98.4,95.8) &(100.0,98.8) &(99.9,99.5) \\
v=5.0 (PRO) &96.6 & 95.2 &84.0 & 84.0 &93.0 &98.5 &99.2 & 93.8 &92.4 &81.9 &96.1 &94.4 \\
 \midrule
v=6.0 (AUROC) &(99.9,98.7) & (100.0,99.5) &(96.5,94.9) &(98.1,96.8) & (99.0,96.8) &(99.2,98.7) &(99.2,98.5) & (99.8,93.0) &(99.8,96.9) &(97.5,96.4) &(100.0,98.6)  &(99.9,99.5) \\
v=6.0 (PRO) &96.4 & 95.2 &65.2  &85.1 &93.9 &98.3 &99.2 &93.7 &92.3 &85.5 &95.8 &94.8 \\
\midrule
v=7.0 (AUROC) &(99.9,98.7) & (100.0,99.5) &(96.0,94.5) &(98.1,96.5) &(99.0,96.9) &(99.2,98.7) &(99.2,98.4) &(100,93.4) &(99.7,97.4) &(97.5,96.3) &(100.0,98.5) &(100.0,99.5) \\
v=7.0 (PRO) &96.1 & 95.0 &64.2 &85.1& 94.2 &98.5 &99.2 &93.3 &93.3 &85.7 &95.5 &94.7 \\
\midrule
v=8.0 (AUROC) &(99.9,98.7) & (100.0,99.4) &(95.4,94.1) &(98.1,96.2) &(98.9,97.0) &(99.1,98.6) &(99.2,98.4) &(99.8,91.2) &(99.7,97.3) &(98.4,95.6) &(100.0,98.4) &(100.0,99.5) \\ 
v=8.0 (PRO) &96.5 & 94.9 &63.4 &85.0 &93.4 &98.4 &99.3 &93.3 &93.5 &81.5 &95.3 & 94.2 \\
\bottomrule
\end{tabular}}
\label{Table: vparameter-visa}
\end{table*}

%% file: tables/W-WO-VisA.tex
\begin{table*}[ht]
\centering
\caption{Impact of the conditioning on VisA \cite{zou2022spot} only when compared in pixel-wise distance. AUROC metric is in the format of (Image AUROC, Pixel AUROC)} 
\resizebox{\textwidth}{!}{\begin{tabular}{cccccccccccccc}
\toprule
\textbf{Categories}  &Candle & Capsules &Cashew &Chewing gum &Fryum &Macaroni1 &Macaroni2 &PCB1 &PCB2 &PCB3 &PCB4 &Pipe fryum &Avg \\
\midrule
 W/O conditioning - AURO &(79.6,88.1) & (80.5,99.4) &(87.4,63.7) &(92.5,70.6) & (85.9,94.5) & (73.8,92.7) &(69.3,94.3) &(90.8,88.7) &(98.6,97.1) &(99.7,97.1) &(98.9,93.2) &(86.5,77.4) &(87.0, 88.1)\\
 
W/O conditioning - PRO &82.4 &94.1 &39.8 &53.2 &93.1 &95.3 &97.4 &93.4 &94.3 &95.9 &78.1 &74.8 &82.7 \\
\hline
W conditioning - AUROC &(91.9,95.9) &  (91.2,99.7) &(87.4,63.7) &(97.2,85.3) & (94.9,95.4) &(97.2,99.5) &(80.4,98.2) &(95.5,69.2) &(98.8,95.4) & (99.0,95.5) &(98.9,96.1) & (97.2,97.7) & (94.1, 91.0)\\
W conditioning - PRO &94.3 &97.7 &39.8 &77.1 &93.6 &99.7 &99.4 &88.4 &92.5 &94.5 &90.3 &97.2 &88.7 \\
\bottomrule
\end{tabular}}
\label{Table: W-WO conditioning VisA}
\end{table*}

\begin{table*}[ht]
\centering
\caption{Impact of the domain adaptation on VisA \cite{zou2022spot} only when compared in feature-wise distance. AUROC metric is in the format of (ImageAUROC, PixelAUROC)} 
\resizebox{\textwidth}{!}{\begin{tabular}{cccccccccccccc}
\toprule
\textbf{Categories}  &Candle & Capsules &Cashew &Chewing gum &Fryum &Macaroni1 &Macaroni2 &PCB1 &PCB2 &PCB3 &PCB4 &Pipe fryum &Avg \\
\midrule

 W/O conditioning - AURO &(75.1,87.7) & (54.5,86.8) &(90.4,97.1) &(96.5,95.6) &(88.3,78.4) &(61.5,69.9) &(55.7,78.5) &(55.1,71.2) &(53.1,47.0) &(59.2,37.7) &(21.0,59.7) &(50.4,29.5)& (63.4,69.9)\\
W/O conditioning - PRO  &65.3 & 45.5 &83.4 &72.9 & 55.1 &22.1 &39.2 &4.0 &4.2 &0.4 &8.4 &13.0 &34.5\\
\hline

W conditioning - AUROC  &(90.0,93.4) & (90.4,97.6) &(90.4,97.1) &(96.5,95.6) &(96.6,72.3) &(85.8,98.1) &(90.4,98.4) &(88.1,97.8) &(85.8,94.1) &(78.4,91.8) &(97.7,98.4) &(76.0,69.5) &(95.6,92.0)\\
W conditioning - PRO  &79.6 &82.1 &83.4 &72.9 &60.9 &95.4 &95.5 &86.7 &80.7 &63.1 &86.7 &59.8 &78.9 \\
\bottomrule
\end{tabular}}
\label{Table: W-WO domain adaptatoin VisA}
\end{table*} 

%% file: tables/feature_extractor.tex
\begin{table*}[ht]
\centering
\caption{Performance of various feature extractors on MVTec \cite{bergmann2019mvtec}, in the format (Image AUROC, Pixel AUROC)} 
\resizebox{\textwidth}{!}{\begin{tabular}{ccccccccccccccccc}
\toprule
 \textbf{Categories} &Carpet &Grid &Leather & Tile &Wood &Bottle &Cable &Capsule &Hazelnut & Metal nut &Pill &Screw &Toothbrush &Transistor &Zipper &Avg\\
 \midrule
\textbf{ResNet-50}&
(96.7,98.4) &(100,98.0) &(99.9,97.0) & (100,97.5) &(85.4,89.2) &(98.6,97.6) &(93.4,97.3) & (69.1,73.4) &(78.8,92.6) & (91.2,96.2) &(57.8,82.6) &(53.2,60.6) &(75.6,95.8) &(99.9,93.0) &(95.2,88.4) &(86.3,90.5)\\
\textbf{ResNet-50+DA} &(96.7,98.4) &(100,98.8) &(100,99.0) & (100,97.5) &(84.7,86.9) &(99.4,98.0) &(98.5,98.0) &(98.8,94.0) &(99.4,95.9) & (99.4,97.4) &(97.1,98.5) &(94.1,98.7) & (93.6,97.2) &(100,93.8) &(98.7,97.8) &(97.4,96.7)\\
\textbf{ResNet-50+DA+pixel} &(97.6,98.3) &(100,99.1) &(100,99.3) & (100,97.8) &(94.3,91.0) &(99.8,98.4) &(99.4,98.0) &(99.2,95.2) &(100,97.9) & (100,97.7) &(99.4,99.2) &(99.2,99.3) &(100,98.2) &(100,93.3) &(99.9,98.0) &(99.3,97.4)\\
\midrule
\textbf{WideResNet-50} &(99.2,98.8) &(100,98.2) &(100,97.3) & (100,97.5) &(87.6,91.7) &(98.9,97.8) &(96.9,97.5) &(71.1,71.4) & (84.9,92.8) & (89.5,95.6) &(60.8,77.8) &(58.6,53.5) &(77.2,96.0) &(99.9,94.2) &(94.3,85.6) & (88.0, 89.7)\\
\textbf{WideResNet-50+DA} & (99.2,98.8) &(100,99.0) &(100,99.3) & (100,97.5) &(95.5,93.2) &(99.8,98.2) &(98.5,98.0) &(98.3,93.5) &(98.6,95.6) & (99.4,97.8) &(93.6,97.8) &(95.3,99.1) &(93.3,97.8) &(100,93.8) &(99.8,96.6) &(98.1,97.1)\\
\textbf{WideResNet-50+DA+pixel} &(99.6,98.8) &(100,99.3) &(100,99.4) & (100,97.8) &(99.7,94.4) &(100,98.5) &(99.7,98.0) & (98.7,95.1) &(99.7,97.6) &(99.9,98.0) &(97.5,98.9) &(98.2,99.4) &(100,98.5) &(100,94.8) &(100,98.0) & 
(99.5,91.2)\\
\bottomrule
\end{tabular}}
\label{Table:feature-extractor-MVTec}
\end{table*}

\begin{table*}[ht]
\centering
\caption{Performance of various feature extractors on VisA \cite{zou2022spot}, in the format (Image AUROC, Pixel AUROC)} 
\resizebox{\textwidth}{!}{\begin{tabular}{ccccccccccccccc}
\toprule
\textbf{Categories} &Candle & Capsules &Cashew &Chewing gum &Fryum &Macaroni1 &Macaroni2 &PCB1 &PCB2 &PCB3 &PCB4 &Pipe fryum &Avg\\
 \midrule
\textbf{ResNet-50} &(69.4,92.6) & (63.8,91.9) &(90.4,91.5) &(94.9,94.0) &(93.9,80.8) &(76.3,75.7) &(56.9,68.7) &(46.9,68.4) &(50.7,53.7) &(60.6,45.8) & (31.9,66.3) &(47.4,28.6) &(65.3,71.5)\\
\textbf{ResNet-50+DA} &(83.7,94.8) & (92.0,98.8) &(90.4,91.5) &(94.9,94.0)  &(95.7,75.8) &(87.5,93.8) &(72.5,96.3) &(84.4,95.7) &(86.3,90.7) &(79.8,77.5) &(98.6,99.0) &(76.0,67.0) &(86.8,89.6)\\
\textbf{ResNet-50+DA+pixel} &(99.5,98.4) & (99.9,99.4) &(93.5,95.2) &(97.7,94.5) &(100,93.6) &(97.4,94.9) &(83.9,94.6) &(100,92.7) &(95.4,96.5) &(97.7,97.1) &(100,98.8) &(99.1,99.4) &
(97.0, 96.3)\\
\midrule
\textbf{WideResNet-50} & 
(70.8,92.4) & (64.9,92.6) &(91.9,93.0) &(95.1,94.6) &(88.4,77.8) &(69.5,72.8) & (57.8,65.2) &(61.5,68.8) &(56.1,45.8) &(56.0,33.7) &(26.6,65.2) &(49.6,35.7) &
(65.7,69.8)\\
\textbf{WideResNet-50+DA} &(84.0,95.3)& (89.2,98.9) &(91.9,93.0) &(95.1,94.6) &(96.8,80.9) &(93.3,97.9) &(84.3,98.7) &(86.7,97.8) &(86.8,94.2) &(82.5,88.4) &(99.1,98.9) &(80.5,73.6) &(86.7,92.7 )\\
\textbf{WideResNet-50+DA+pixel} &(99.7,98.2) & (99.9,99.6) &(97.0,94.2) &(99.8,95.8) &(100,93.2) &(99.4,97.4) &(87.6,90.1) &(99.9,93.5) &(98.0,95.2) &(89.2,93.9) &(100,98.7) &(99.9,99.4) &
(97.5, 95.8)\\
\midrule
\textbf{WideResNet-101} &(75.1,87.7) & (54.5,86.8) &(90.4,97.1) &(96.5,95.6) &(88.3,78.4) &(61.5,69.9) &(55.7,78.5) &(55.1,71.2) &(53.1,47.0) &(59.2,37.7) &(21.0,59.7) &(50.4,29.5)& (63.4,69.9)\\
\textbf{WideResNet-101+DA} &(91.9,95.9) &  (91.2,99.7) &(87.4,63.7) &(97.2,85.3) & (94.9,95.4) &(97.2,99.5) &(80.4,98.2) &(95.5,69.2) &(98.8,95.4) & (99.0,95.5) &(98.9,96.1) & (97.2,97.7) & (94.1, 91.0)\\
\textbf{WideResNet-101+DA+pixel} &(99.9,98.7) & (100,99.5) &(94.5,97.4) &(98.1,96.5) &(98.9,96.4) &(99.2,98.7) &(99.3, 98.4) &(100,93.4) &(99.7,97.4) &(97.2,96.3) &(100,98.5) &(100,99.5) &(98.9,97.6)\\
\bottomrule
\end{tabular}}
\label{Table:feature-extractor-visA}
\end{table*}

%% file: tables/parameter-lambda.tex
\begin{table*}[ht]
\centering
\caption{Role of \({\lambda}_{DL}\) in improving results on MVTec \cite{bergmann2019mvtec}. Format (ImageAUROC, PixelAUROC)} 
\resizebox{\textwidth}{!}{\begin{tabular}{ccccccccccccccccc}
\toprule
 \textbf{Categories} &Carpet &Grid &Leather & Tile &Wood &Bottle &Cable &Capsule &Hazelnut & Metal nut &Pill &Screw &Toothbrush &Transistor &Zipper &Avg\\
 \midrule
WO   &(99.3,98.7) &(100,98.9) &(100,97.9) & (99.7,97.1) &(92.1,94.8) &(100,98.5) &(99.4,98.1) &(85.8,87.4) &(96.6,97.4) & (96.8,98.0) &(75.1,90.3) & (74.7,88.3) &(100,98.6) & (100,95.3) &(99.7,95.1) &(94.6,95.6) \\
\({\lambda}_{DL}=0\)   &(99.3,98.7) & (100,99.4) &(100,99.1) & (100,97.2) &(96.7,89.1) &(100,98.6) &(99.4,98.1) &(97.8,93.5) &(99.4,98.8) & (99.0,98.2) &(99.5,98.1) & (96.9,98.6) &(100,98.7) & (100,95.3) & (99.9,95.6) &(99.2,97.1) \\
\({\lambda}_{DL}=0.1\)   &(99.3,98.7) & (100,99.4) &(100,99.4) & (100,98.2) &(100,95.3) &(100,98.7) &(99.4,98.1) &(99.4,95.7) &(100,98.3) &(100,98.9) &(100,99.1) &(99.0,99.3) &(100,98.7) &(100,95.3) & (100,98.2) &(99.8,98.1) \\
 \({\lambda}_{DL}=0.2\)  &(99.3,98.7) &(100,99.3) &(100,99.4) & (100,98.2) &(100,95.4) &(100,98.6) &(99.4,98.3) & (99.1,95.2) &(100,98.3) & (100,98.7) &(99.7,99.0) &(99.0,99.1) &(100,98.7) &(100,95.3)  &(100,98.2) &(99.8, 98.0)\\
\bottomrule
\end{tabular}}
\label{Table: lambda-DL-MVTec}
\end{table*}

% \begin{table*}[ht]
% \centering
% \caption{Sensitivity of conditioning parameter \(w\) on MVTec \cite{bergmann2019mvtec} only when compared in pixel-wise distance. Format (ImageAUROC, PixelAUROC)} 
% \resizebox{\textwidth}{!}{\begin{tabular}{cccccccccccccccc}
% \toprule
%  \textbf{Categories}&Candle & Capsules &Cashew &Chewing gum &Fryum &Macaroni1 &Macaroni2 &PCB1 &PCB2 &PCB3 &PCB4 &Pipe fryum \\
%  \midrule
% WO  &Candle & Capsules &Cashew &Chewing gum &(98.0,95.3) &Macaroni1 &Macaroni2 &PCB1 &PCB2 &PCB3 &PCB4 &Pipe fryum  \\
% \({\lambda}_{DL}=0\)  &Candle & Capsules &Cashew &Chewing gum &(99.9,93.2) &Macaroni1 &Macaroni2 &PCB1 &PCB2 &PCB3 &PCB4 &Pipe fryum  \\
% \({\lambda}_{DL}=0.1\)  &Candle & Capsules &Cashew &Chewing gum &Fryum &Macaroni1 &Macaroni2 &PCB1 &PCB2 &PCB3 &PCB4 &Pipe fryum \\
% \bottomrule
% \end{tabular}}
% \label{Table: feature_extractor}
% \end{table*} 